\documentclass{article} 
\usepackage{iclr2027_conference,times}

\usepackage{amsmath,amsfonts,bm}

\def\eqref#1{equation~\ref{#1}}

\def\1{\bm{1}}

\DeclareMathAlphabet{\mathsfit}{\encodingdefault}{\sfdefault}{m}{sl}
\SetMathAlphabet{\mathsfit}{bold}{\encodingdefault}{\sfdefault}{bx}{n}

\usepackage{url}

\definecolor{emreblue}{RGB}{61,150,209}
\usepackage{wrapfig}
\usepackage[colorlinks=true, citecolor=emreblue, linkcolor=emreblue, urlcolor=emreblue]{hyperref}
\usepackage{cleveref}
\usepackage{tocbibind} 
\usepackage{titletoc}  
\usepackage[utf8]{inputenc}
\usepackage{microtype}
\usepackage{graphicx}
\usepackage{multirow}
\usepackage{url}
\usepackage{times}
\usepackage{amsmath}
\usepackage{amssymb}
\usepackage{booktabs}
\usepackage{xcolor}     
\usepackage{natbib}
\definecolor{mydarkblue}{rgb}{0,0.08,0.45}
\usepackage{fancyhdr}
\usepackage{wrapfig}
\usepackage[textwidth=2.5cm]{todonotes}
\usepackage{tabularx}

\usepackage{enumitem}
\newenvironment{itemize*}%
 {\leftmargini=20pt\begin{itemize}%
  \setlength{\itemsep}{3pt}%
  \setlength{\parskip}{0pt}%
  }%
 {\end{itemize}} 
\newenvironment{enumerate*}%
 {\begin{enumerate}%
  \setlength{\itemsep}{0pt}%
  \setlength{\parskip}{0pt}}%
 {\end{enumerate}}

\definecolor{mylightgray}{RGB}{240,240,240}
\newcommand{\highlightpink}[1]{%
\tikz[baseline=(char.base)]{%
\node[%
shape=rectangle,%
rounded corners=2pt,%
draw=mylightgray,%
fill=mylightgray,%
inner sep=2pt,%
text width=0.7cm,%
align=center,%
text height=1.3ex,%
text depth=.1ex%
] (char) {#1};%
}%
}

\usepackage[most]{tcolorbox}
\newtcolorbox{guidance}{
  colback=mylightgray,
  colframe=mylightgray,
  boxrule=0pt,
  arc=3pt,
  left=6pt, right=6pt, top=4pt, bottom=4pt,
  before skip=4pt, after skip=4pt,
}

\definecolor{mylightblue}{HTML}{EAF1FA}

\newtcolorbox{discussion}{
  colback=mylightblue,
  colframe=mylightblue,
  boxrule=0pt,
  arc=3pt,
  left=6pt, right=6pt, top=4pt, bottom=4pt,
  before skip=4pt, after skip=4pt,
}

\usepackage{pifont}

\definecolor{okgreen}{HTML}{2E8B57}
\definecolor{nored}{HTML}{C0392B}
\definecolor{amber}{HTML}{E0A800}

\definecolor{opdblue}{RGB}{34,87,143}
\definecolor{rlvrgreen}{RGB}{28,107,82}
\definecolor{sftgold}{RGB}{181,128,31}
\definecolor{gapred}{RGB}{176,48,48}

\newcommand{\sflogo}{\raisebox{-0.15em}{\includegraphics[height=1.0em]{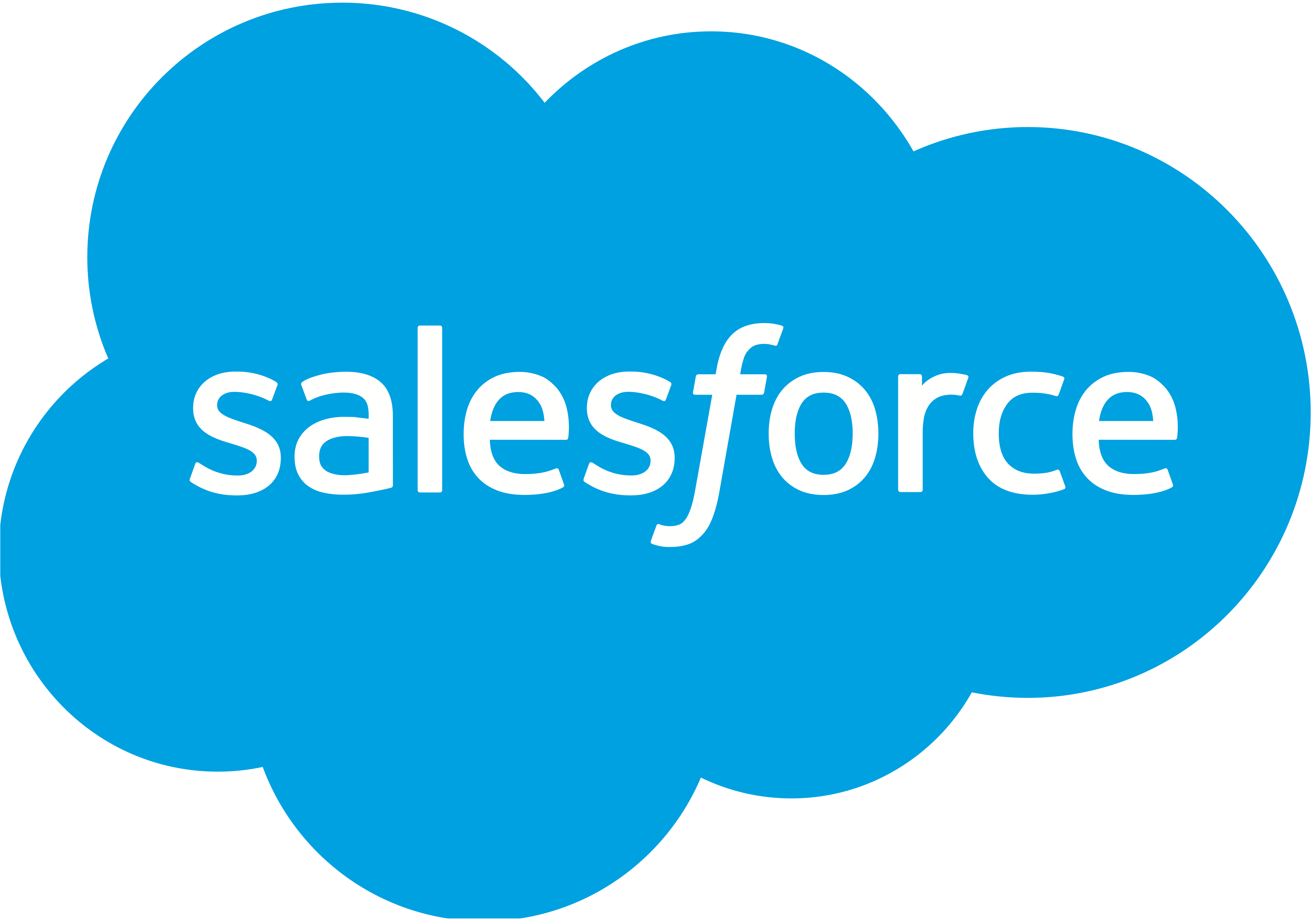}}}
\newcommand{\uiuclogo}{\raisebox{-0.15em}{\includegraphics[height=1.0em]{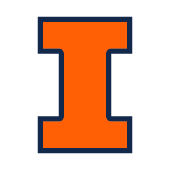}}}

\title{Understanding the Synergy between\\SFT, RLVR, and OPD in LLM Post-Training}

\author{Emre Can Acikgoz$^{1,2,\dagger,\ddagger}$, \,Yang Li$^{1,\dagger}$, \,Zeyu Leo Liu$^{1}$, \,Srijan Bansal$^{1}$, \\
\textbf{Dilek Hakkani-T\"ur$^{2}$, \,Shafiq Joty$^{1}$, \,Semih Yavuz$^{1}$} \\
$^{1}$ \sflogo\,Salesforce AI Research, $^{2}$\uiuclogo UIUC
}
\iclrfinalcopy 
\begin{document}

\maketitle
\begingroup
\makeatletter
\renewcommand\@makefntext[1]{\noindent#1}
\makeatother
\renewcommand\thefootnote{}
\footnotetext{$^{\dagger}$Equal contribution. \, $^{\ddagger}$Work done during internship at Salesforce AI Research.}
\endgroup

\begin{abstract}
Modern LLM post-training composes supervised fine-tuning (SFT), reinforcement learning with verifiable rewards (RLVR), and on-policy distillation (OPD) into multi-stage pipelines, yet these stages are typically designed and evaluated in isolation.
We show that this composition is consequential: a stage that improves the current model can make the next stage less effective. 
Through controlled experiments with Qwen3 models on math and science reasoning, we first characterize OPD across nine student–teacher pairs spanning 2$\times$ to 53$\times$ parameter ratios and show that OPD effectiveness depends on student--teacher compatibility rather than teacher scale alone. 
The surrounding stages of OPD reshape this compatibility in three ways:  
(1) A brief SFT warm-up improves subsequent OPD, while an RLVR-strengthened student regresses under distillation from the same teacher.
(2) Adapting the teacher with RLVR raises downstream OPD accuracy in proportion to the capability it adds.
Following these two interventions, we find that combining teacher adaptation and student warm-up alone raise average OPD accuracy from 29.2\% to 43.8\% (\highlightpink{\textbf{50\%}} relative improvement) after the same number of distillation steps, with additional preparatory training.
(3) At comparable accuracy, OPD leaves a stronger initialization for downstream RLVR than SFT, with a gap that widens as RL compute scales. 
Our results suggest that each post-training stage should be chosen not only for the capability it adds, but for the \textit{learning interface} it creates for the next stage.
\end{abstract}


\vspace{-5mm}

\begin{figure*}[h!]
\centering
\includegraphics[width=\textwidth]{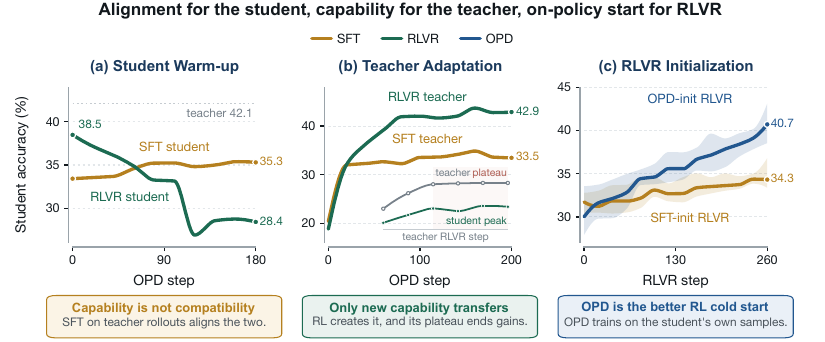}
\vspace{-6mm}
\caption{
\textbf{The outcome of each post-training stage depends on the learning state it inherits.}
\textbf{(a)}~An RLVR-warmed student (\textcolor{rlvrgreen}{green}) starts stronger but regresses under OPD, while a briefly SFT-warmed student (\textcolor{sftgold}{gold}) improves steadily from the same teacher.
\textbf{(b)}~An RLVR-adapted teacher transfers substantially more than a SFT-adapted teacher of the same origin. 
\textit{Inset:} teacher accuracy along its RLVR training (\textcolor{gray}{gray}) and final OPD accuracy of a student distilled from the teacher at each step (\textcolor{rlvrgreen}{green}); both plateau together, so student gains stop when the teacher's do.
\textbf{(c)}~At comparable starting accuracy, OPD-initialized models (\textcolor{opdblue}{blue}) reach higher final accuracy under subsequent RLVR than SFT-initialized models, and the gap widens with RL compute.
}
\vspace{-5mm}
\label{fig:abs}
\end{figure*}


\section{Introduction}
\label{sec:intro}

Recent advances in post-training have substantially improved the reasoning capabilities of large language models (LLMs)~\citep{guo2025deepseekr1, lambert2025tulu, yang2025qwen3, xiao2026mimo, zeng2026glm}.
Three post-training objectives drive this progress: supervised fine-tuning (SFT) on expert demonstrations~\citep{ouyang2022instructgpt, wei2022flan}, reinforcement learning with verifiable rewards (RLVR)~\citep{lambert2025tulu, shao2024deepseekmath} with sparse outcome-level feedback on model-generated trajectories, and on-policy distillation (OPD)~\citep{lu2025onpolicydistillation, yang2025qwen3} with dense token-level teacher supervision on the student's own trajectories. 
In practice these stages are rarely applied in isolation: SFT, RLVR, and OPD are composed sequentially, each reshaping the policy that the next stage receives. 
Yet, despite the rapid adoption of multi-stage post-training pipelines, how these stages interact and in what order they should be composed remains poorly understood. 
We show that this composition problem is consequential: \textit{the effect of a post-training stage depends on the learning state created by the stage before it}, and a stage that improves the current model can simultaneously make the subsequent stage less effective.

We begin by characterizing what determines OPD effectiveness in isolation~(\S\ref{sec:dynamics}). 
Consistent with concurrent findings~\citep{li2026rethinking, zhu2026many, wang2026demystifying}, we observe across nine Qwen3 student--teacher pairs spanning $2\times$ to $53\times$ parameter ratios that stronger teachers do not necessarily transfer better under OPD. 
Moreover, reverse KL continues to decrease after accuracy has already peaked, so closer teacher matching does not imply better task performance either. 
We show that \emph{reinforceable fraction} can be a complementary diagnostic that captures student--teacher compatibility where scalar divergence does not. 
Together, these results suggest that OPD effectiveness depends on the student--teacher pairing at initialization. 
Since SFT, RLVR, and OPD each reshape that pairing, their ordering cannot be treated as arbitrary.

We call this the \textit{composition problem}: choosing the order, duration, and handoff points of SFT, RLVR, and OPD in a post-training pipeline~(\S\ref{sec:composition}). 
Current frontier pipelines compose these stages in different ways: some warm-up the student with SFT and then apply RLVR~\citep{guo2025deepseekr1, shao2024deepseekmath}, while recent work inserts OPD before RLVR~\citep{xiao2026mimo, zeng2026glm}. 
On the teacher side, some pipelines strengthen the teacher with RLVR before distillation~\citep{yang2025qwen3}, and others interleave domain-specific RL with OPD to consolidate multiple specializations into a single model~\citep{yang2026nemotron}. 
Yet these choices are rarely compared under a shared experimental protocol.
Does improving the student with SFT or RLVR help subsequent OPD, and much SFT warm-up is useful before it begins to hinder distillation~\citep{li2026rethinking, zhu2026many}? 
How do teacher accuracy gains from RLVR translate into student gains, compared with SFT on verified solutions~\citep{zelikman2022star, zhu2026many}? 
And when OPD is followed by RLVR, does distillation leave a better initialization than SFT~\citep{aphale2026sft, shen2026stage}?


To answer these questions, we design controlled experiments that vary the student state, teacher state, and order of training stages. 
Our main study uses a Qwen3-4B-Base student and a Qwen3-14B teacher, evaluated on six math and one science benchmark.
Within each comparison, we keep the surrounding data, optimization setup, and evaluation protocol fixed so that the impact of each stage can be isolated.
From these experiments, we derive the following findings: 
\begin{itemize}[topsep=0pt, leftmargin=10pt, itemsep=0pt]
    \item Increasing teacher scale does not consistently improve distillation across student sizes, model families, objectives, and training distributions, highlighting the importance of student--teacher compatibility.
    \item Brief SFT warm-up improves distillability while bringing the student closer to the teacher, whereas a stronger RLVR-prepared student can regress under distillation from the same teacher (Figure~\ref{fig:abs}a).
    \item Teacher adaptation with RLVR improves downstream OPD as teacher accuracy rises, with diminishing returns once teacher performance plateaus (Figure~\ref{fig:abs}b).
    \item At comparable starting accuracy, OPD provides a more effective initialization than SFT for subsequent RLVR and aggregate policy entropy alone does not account for this advantage (Figure~\ref{fig:abs}c).
\end{itemize}

Across all experiments, the same pattern emerges: the outcome of a stage depends on the state it inherits, so a stage should be judged not only by the capability it adds but also by the state it hands to the next stage, which we call its \emph{learning interface}.
In addition, these findings motivate a simple composition recipe: \textbf{adapt the teacher with RLVR, briefly warm the student with SFT, and then distill}. 
Combining teacher adaptation and student warm-up raises average final OPD accuracy from $29.2\%$ to $43.8\%$ ($50\%$ relative improvement) at the same OPD step budget, with additional preparatory training.\looseness-1

We summarize the overall contributions of our paper as follows:
\begin{itemize}[topsep=0pt, leftmargin=10pt, itemsep=0pt]
    \item We characterize OPD across model scales, families, objectives, and training distributions under a shared protocol, confirming that effective distillation depends on student--teacher compatibility rather than teacher scale alone~(\S\ref{sec:dynamics}).
    \item We present a systematic controlled study of how SFT, RLVR, and OPD interact, showing that student capability and distillability can move in opposite directions, that teacher adaptation transfers in proportion to the capability it adds, and that OPD provides a stronger RLVR initialization than SFT at comparable accuracy~(\S\ref{sec:composition}).
    \item We derive an empirically grounded composition recipe, where we first adapt teacher with RLVR, then briefly warm-up the student with SFT, and then transition to OPD. This recipe demonstrate significant improvement within the settings studied~(\S\ref{sec:composition_rule}).
\end{itemize}
By studying these interactions in open-weight models at scale, we aim to make LLM post-training dynamics more accessible to the research community and to inform the design of future pipelines.

\vspace{-6pt}
\section{Preliminaries}
\label{sec:background}
\vspace{-4pt}

We define the post-training objectives studied throughout the paper and the quantities we use to characterize student--teacher transfer.
Appendix~\ref{app:objectives} gives full objective details, Appendix~\ref{app:metrics} provides metric definitions, aggregation, and derivations, and Appendix~\ref{app:gap} extends the related-work discussion.\looseness-1

\textbf{Post-Training Objectives.~}
Given a labeled dataset $\mathcal{D}=\{(x^{(i)},y^{(i)})\}$ of trajectories, \textbf{SFT} minimizes token-level cross entropy: $\mathcal{L}_{\text{SFT}}(\theta) = -\mathbb{E}_{(x,y)\sim\mathcal{D}}\bigl[\sum_{t} \log \pi_{S_\theta}(y_t \mid x, y_{<t})\bigr]$.
The supervision here is off-policy: prefixes and targets are fixed regardless of what the evolving student generates.
\textbf{RLVR}  optimizes the student on its own rollouts using a scalar outcome reward $r(x,y) \in \{0,1\}$. 
We use GRPO~\citep{shao2024deepseekmath}, which samples $G$ responses per prompt and normalizes within the group: $\hat{A}^{\text{RLVR}}_i = ({r(x, y^{(i)}) - \mu_{\text{group}}})/{\sigma_{\text{group}}}$.
The supervision is on-policy but sparse: one outcome reward per self-generated trajectory, shared by every token.
\textbf{OPD} also trains on student-generated trajectories with dense teacher supervision at every prefix~\citep{gu2023minillm, agarwal2024policy, lu2025onpolicydistillation}. 
Our default formulation uses a sampled-token policy-gradient surrogate for token-level reverse-KL matching~\citep{lu2025onpolicydistillation}. 
The objective minimizes reverse KL between the student $p_t$ and teacher $q_t$ at each prefix: $\mathcal{L}_{\text{OPD}} = \mathbb{E}_{x, y\sim\pi_S}\bigl[\frac{1}{|y|}\sum_{t} D_{\text{KL}}(p_t \| q_t)\bigr]$. 
Our default implementation uses a sampled-token policy-gradient estimator optimized with the same clipped importance-ratio objective used for RLVR. 
We also evaluate a dense top-$k$ forward-KL objective for robustness (Appendix~\ref{app:objectives}).

The three methods differ along two axes: SFT provides dense supervision on fixed trajectories, RLVR provides sparse outcome supervision on student trajectories, and OPD provides dense teacher supervision on student trajectories.

\textbf{Measuring Transfer.~}
We characterize student--teacher transfer with three quantities (formal definitions in Appendix~\ref{app:metrics}). 
Let $A_S$, $A_{S'}$, and $A_T$ be the accuracies of the original student $S$, the distilled student $S'$, and the teacher $T$. 
\textbf{Distillation effectiveness} $\text{DE} = (A_{S'} - A_S)/(A_T - A_S)$ is the fraction of the accuracy gap recovered. 
The \textbf{reinforceable fraction} $\rho_0^+$ is the share of student-sampled tokens receiving a positive teacher signal at initialization.
It captures how often the teacher reinforces the student's own behavior, providing a compatibility signal that teacher accuracy alone cannot reveal.
\textbf{Normalized reverse KL} $\kappa_s = \widehat{\text{rKL}}_s / \widehat{\text{rKL}}_0$, where $\widehat{\text{rKL}}_s$ estimates $\mathrm{KL}(\pi_{S,s} \,\|\, \pi_T)$ on student samples at step $s$ and tracks on-policy divergence during training. 
Together, these capture outcome, initialization-time compatibility, and training dynamics.

\textbf{From Individual Stages to Composition.~}
Several concurrent studies have characterized OPD dynamics in isolation. \citet{li2026rethinking} identify thinking-pattern consistency and teacher-knowledge novelty as conditions for successful transfer. \citet{wang2026demystifying} characterize OPD as an exploration catalyst and attribute failures to signal corruption. \citet{zhu2026many} document pitfalls of the reverse-KL approximation. On adjacent stages, prior work has shown that SFT cold starts improve OPD~\citep{li2026rethinking, zhu2026many}, that RLVR teacher adaptation aids downstream distillation~\citep{zhu2026many}, and that pre-RL entropy predicts post-RLVR outcomes~\citep{aphale2026sft}. What remains less understood is how these stages \emph{compose}: how changing one stage alters the learning conditions faced by the next. We address this gap by first characterizing OPD under a controlled protocol~(\S\ref{sec:dynamics}), then studying how SFT, RLVR, and OPD reshape student--teacher compatibility when composed~(\S\ref{sec:composition}). A detailed discussion of prior work is in Appendix~\ref{app:gap}.

\section{OPD Dynamics at Scale}
\label{sec:dynamics}

\vspace{-4mm}
\begin{figure*}[h!]
\centering
\includegraphics[width=\textwidth]{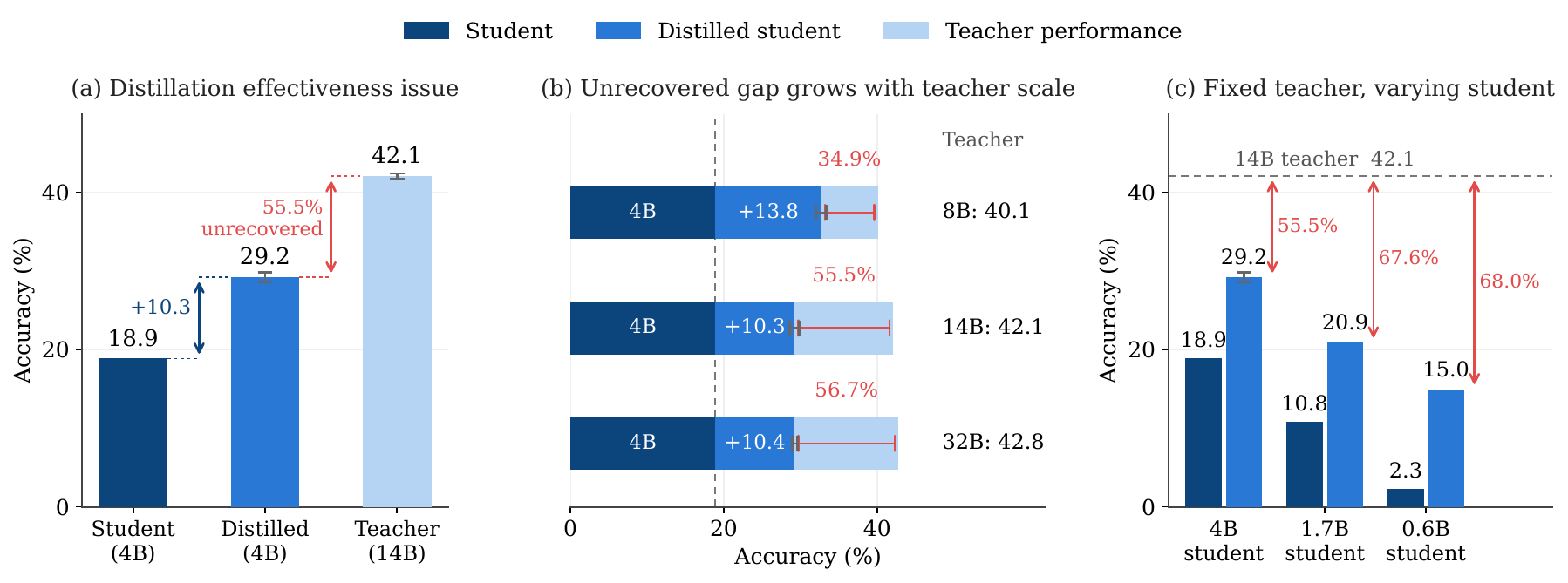}
\vspace{-7mm}
\caption{
\textbf{A larger teacher does not guarantee better transfer.} \textcolor{gapred}{Red} lines show the unrecovered gap, $100 - \text{DE}$ (\%).
\textbf{(a)}~Qwen3-4B-Base gains $10.3$ points from Qwen3-14B but leaves $55.5\%$ of the teacher's advantage unrecovered.
\textbf{(b)}~For the 4B student, DE decreases as teacher size increases from 8B to 32B.
\textbf{(c)}~With the 14B teacher fixed, smaller students recover a smaller share of the gap.
}
\vspace{-1.5mm}
\label{fig:ceiling}
\end{figure*}

\textbf{Setup.} \quad
Before studying how stages compose, we characterize OPD in isolation to establish the baselines and diagnostics that \S\ref{sec:composition} builds on. 
We distill Qwen3-\{0.6B, 1.7B, 4B\}-Base students from Qwen3-\{8B, 14B, 32B\} post-trained teachers (nine pairs, $2\times$--$53\times$ parameter ratios) using DAPO-Math-17K~\citep{yu2026dapo} with sampled-token reverse-KL distillation. 
All runs train for one epoch with 64 prompts and four responses per prompt per batch. 
Unless otherwise stated, we compare at step~200; 4B results average three seeds, and 0.6B and 1.7B use one run per teacher. 
We report \texttt{mean@4} accuracy averaged over AIME~2024, AIME~2025, AMC~2023, MATH-500, Minerva Math, OlympiadBench, and GPQA-Diamond, and quantify transfer using DE~(\S\ref{sec:background}). 
Full configurations are in Appendix~\ref{app:dynamics}.\looseness-1

\textbf{Observation 3.1: Increasing teacher scale does not reliably improve transfer.} \quad Using the largest teacher (Qwen3-32B) yields neither the highest accuracy nor the highest DE for any student in our scaling grid. 
For Qwen3-4B-Base, switching the teacher from 8B to 32B lowers student accuracy from $32.70\%$ to $29.25\%$ and DE from $65.1\%$ to $43.3\%$ (\Cref{fig:ceiling}). 
The preferred teacher also varies across students: Qwen3-8B performs best for both the 4B and 1.7B students, while Qwen3-14B leads for the 0.6B student (Appendix~\ref{app:dynamics}). 
With teacher size held fixed, DE increases with student size for every teacher (\Cref{fig:ceiling}c). 
The complete grid and extended learning curves are in Appendix~\ref{app:dynamics}.

\begin{figure*}[h!]
\centering
\includegraphics[width=0.24\textwidth]{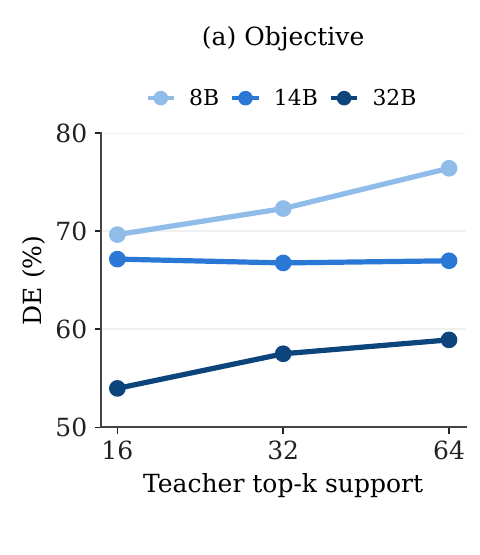}\hfill
\includegraphics[width=0.24\textwidth]{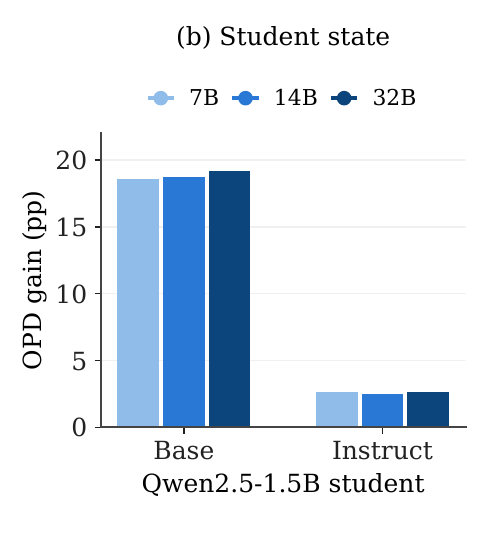}\hfill
\includegraphics[width=0.24\textwidth]{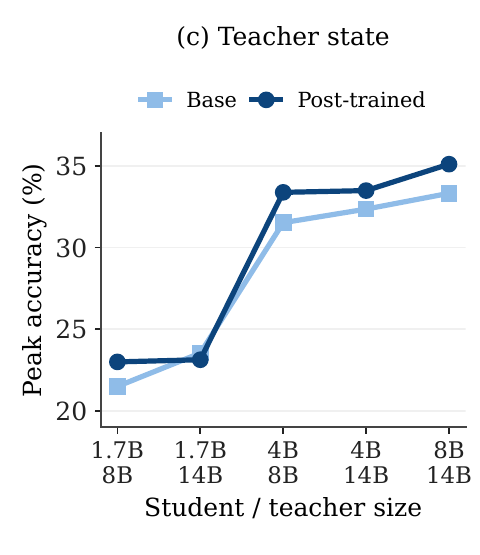}\hfill
\includegraphics[width=0.24\textwidth]{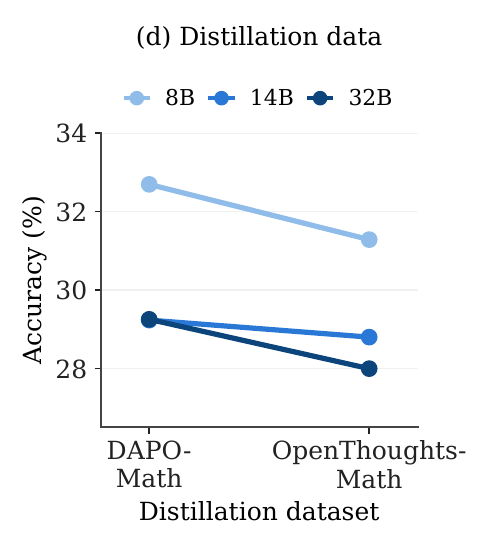}
\vspace{-3mm}
\caption{
\textbf{OPD effectiveness varies across objectives, model states, and training data.}
\textbf{(a)} Under forward KL, DE decreases with teacher size at every top-$k$ support.
\textbf{(b)} Qwen2.5-1.5B base students gain more from OPD than instruct students.
\textbf{(c)} Post-trained teachers improve peak accuracy in four of five matched-size comparisons.
\textbf{(d)} Qwen3-8B gives the highest accuracy on both training datasets for Qwen3-4B-Base.
}
\label{fig:generalization}
\end{figure*}

\textbf{Observation 3.2: OPD effectiveness depends on the distillation setting.}\quad 
The pattern from \textit{Observation~3.1} holds across distillation objectives, student and teacher states, and training data (\Cref{fig:generalization}). 
Under forward KL with top-$k$ support, DE decreases with teacher size at every $k$ (\Cref{fig:generalization}a). 
OPD gains vary substantially with the student's starting state: Qwen2.5-1.5B base students gain $18.54$--$19.14$ points, compared with $2.48$--$2.66$ for instruct students (\Cref{fig:generalization}b). 
Post-trained teachers improve peak accuracy in four of five matched-size comparisons (\Cref{fig:generalization}c).
Qwen3-8B gives the highest accuracy on both DAPO-Math and OpenThoughts-Math for the 4B student (\Cref{fig:generalization}d). 
Full results are in Appendix~\ref{app:dynamics}.

\begin{figure}[htbp]
\centering
\vspace{-4pt}
\includegraphics[width=\linewidth]{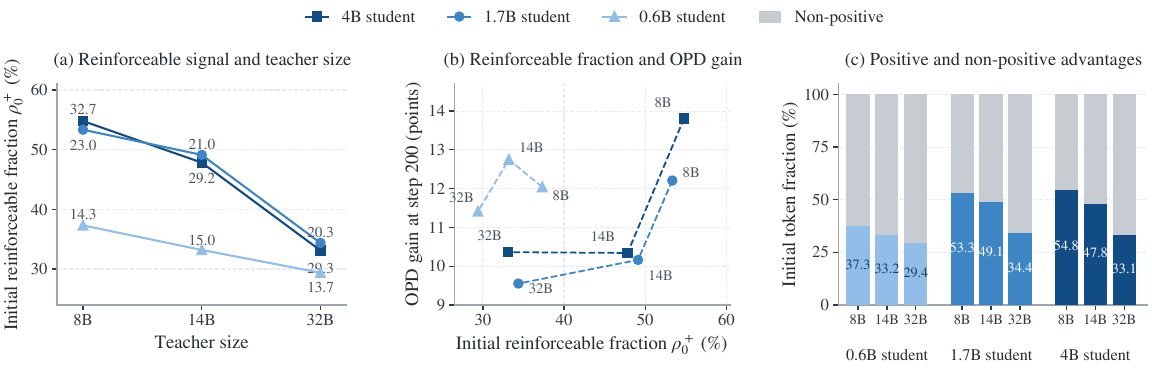}
\vspace{-7mm}
\caption{
\textbf{Reinforceable signal varies across student--teacher pairings.}
\textbf{(a)} Initial $\rho_0^+$ decreases with teacher size for every student, with point labels showing step-200 accuracy.
\textbf{(b)} Selecting maximum $\rho_0^+$ identifies the largest step-200 gain for two of three students.
\textbf{(c)}~Larger teachers leave a greater share of tokens without positive distillation signal.
}
\vspace{-1.5mm}
\label{fig:rho}
\end{figure}

\textbf{Observation 3.3: The reinforceable fraction complements scalar divergence.}\quad Measured before any OPD training, $\rho_0^+$ decreases with teacher size for every student (\Cref{fig:rho}a, \Cref{fig:app_b_distillation_diagnostics}): larger teachers leave more tokens without positive distillation signal (\Cref{fig:rho}c), consistent with their lower OPD gains. 
Reverse KL follows a different pattern: for both 0.6B and 1.7B students, the 32B teacher has the smallest initial rKL but gives the lowest step-200 accuracy. 
Maximum $\rho_0^+$ identifies the best teacher for two of three students at step 200 (\Cref{fig:rho}b), distinguishing productive pairings that scalar divergence does not.

Together, these dynamics suggest that effective OPD depends not on teacher capability or scalar policy distance alone, but on the learning signal induced by the current student--teacher pairing. 
We next ask whether the stages surrounding OPD can deliberately reshape this pairing~(\S\ref{sec:composition}).

\section{Composing Post-Training Stages}
\label{sec:composition}

\S\ref{sec:dynamics} showed that greater teacher scale does not reliably improve OPD and that transfer depends on the current student--teacher pairing.
We now ask how the stages surrounding OPD affect subsequent learning.
SFT, RLVR, and OPD each reshape the policy that the next stage receives, motivating a closer examination of their roles in multistage post-training~\citep{yang2025qwen3, xiao2026mimo, zeng2026glm, yang2026nemotron}.
We study this by changing one component at a time while holding everything else fixed:
\begin{itemize}[topsep=0pt, leftmargin=10pt, itemsep=0pt]
    \item \textbf{Student warm-up} (\S\ref{subsec:student_warmup}): How does preparing the student affect OPD, and do the method and duration matter?
    \item \textbf{Teacher adaptation} (\S\ref{subsec:teacher_adaptation}): Does improving the teacher help, and is it about domain adaptation or more?\looseness-1
    \item \textbf{Stage ordering} (\S\ref{subsec:stage_ordering}): Can OPD be a better initialization for subsequent RLVR than SFT?
\end{itemize}

\textbf{Setup.} \quad 
Unless stated otherwise, all experiments in this section use Qwen3-4B-Base as the student and Qwen3-14B as the teacher with the same optimization recipe and evaluation protocol as in \S\ref{sec:dynamics}. 
Preparatory stages (student warm-up, teacher adaptation, and pre-RLVR SFT or OPD) trained on OpenThoughts-Math~\citep{guha2026openthoughts}, while the final OPD and RLVR stages trained on DAPO-Math~\citep{yu2026dapo}.
Further details are reported in Appendices~\ref{app:student_warmup}--\ref{app:stage_ordering}.

\subsection{Student Warm-Up Before Distillation}
\label{subsec:student_warmup}

\begin{figure*}[t!]
\centering
\vspace{-1mm}
\includegraphics[width=\textwidth]{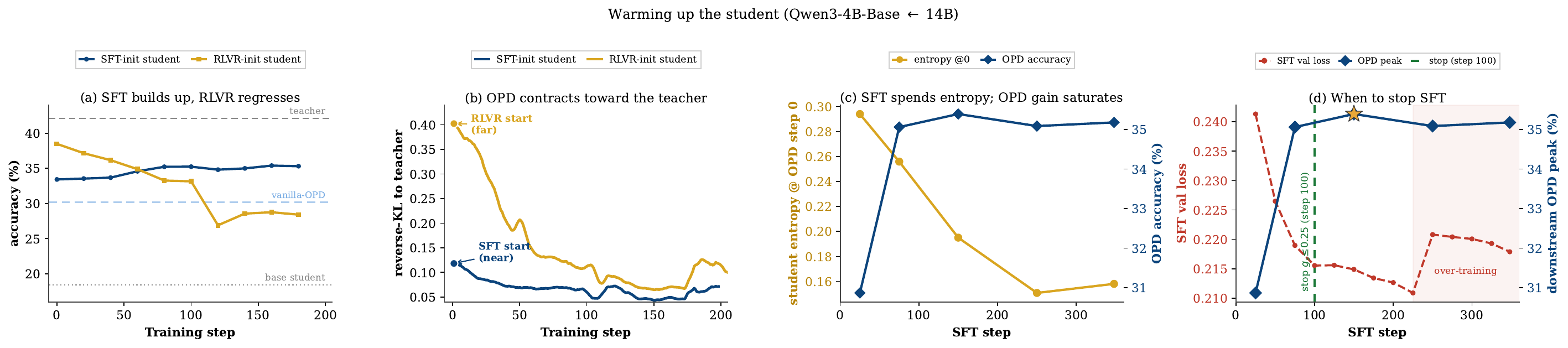}
\vspace{-7mm}
\caption{
\textbf{Student warm-up: the method and the amount both matter.}
\textbf{(a)}~Brief SFT raises the observed OPD performance and reduces the subsequent drop, whereas the RLVR-prepared student regresses under OPD from the same teacher.
\textbf{(b)}~Both students contract toward the teacher in reverse KL---the RLVR-warmed student from much farther away---so the regression occurs despite successful teacher matching.
\textbf{(c)}~As SFT continues, student entropy keeps falling while downstream OPD accuracy saturates after a short warm-up.
\textbf{(d)}~The SFT validation-loss plateau signals when to stop: SFT beyond it no longer improves downstream OPD.
}
\vspace{-4mm}
\label{fig:warmup}
\end{figure*}

If OPD effectiveness depends on the student--teacher pairing at initialization as observed in \S\ref{sec:dynamics}, a natural way to improve distillation is to prepare the student before it begins.
We compare two warm-up methods: (i)~SFT on correct teacher-generated trajectories, which moves the student toward the teacher, and (ii)~RLVR on the target domain, which improves the student independently of the teacher. 
To understand how the amount of SFT shapes the student and the distillation that follows, we sweep the SFT duration across $K \in \{25, 75, 150, 250, 348\}$ steps and track student entropy at OPD initialization.

\textbf{Observation 4.1: Higher student capability does not imply better distillability.} \quad
SFT and RLVR both improve the student before OPD, but have sharply different effects on subsequent distillation (\Cref{fig:warmup}a).
A short SFT warm-up raises the OPD performance from $33.6\%$ to $35.4\%$ and removes the regression that follows the peak in the curve.
The RLVR-warmed student instead enters OPD at $38.5\%$, already within $3.6$ points of the teacher, yet falls to $28.4\%$ by the end of distillation. 
A stronger student is therefore not necessarily a better starting point for distillation.

\textbf{Observation 4.2: Closer teacher matching can accompany student regression.} \quad
Reverse KL decreases during OPD from both SFT and RLVR initializations, even though their accuracy trajectories differ (\Cref{fig:warmup}b).
The SFT checkpoints begin downstream OPD with reported rKL values around $0.10$--$0.14$ (Appendix~\ref{app:student_warmup_diagnostics}), while the illustrated RLVR initialization begins near $0.40$.
For the latter, continued teacher matching does not prevent accuracy from falling.
This extends the distinction in \S\ref{sec:dynamics}: progress in the distillation objective need not imply progress on the task.

\textbf{Observation 4.3: SFT warm-up benefits saturate early.} \quad
Increasing the SFT warm-up initially improves downstream OPD, from $30.9\%$ after $25$ steps to $35.1$--$35.4\%$ after $75$--$150$ steps, and then saturates (Figure~\ref{fig:warmup}c).
Beyond this region, additional SFT provides little improvement in subsequent distillation.
Meanwhile, the reported initial entropy falls from $0.2559$ at $K=100$ to $0.1507$ at $K=250$.
The first SFT validation-loss plateau, near step $100$, lies close to the region where downstream gains saturate (\Cref{fig:warmup}d).
This retrospective agreement suggests a useful signal of diminishing returns, making this plateau a practical indicator of diminishing warm-up returns.

\begin{guidance}
\paragraph{Discussion.}
These experiments separate student \emph{capability} from student \emph{distillability}.
RLVR can produce a substantially stronger student while simultaneously making supervision from the same teacher less useful, whereas a brief SFT warm-up can improve the student--teacher interface even with a much smaller gain in standalone capability.
The contrast is also diagnostic: reverse KL alone does not distinguish the vanilla and RLVR-adapted student states, while the reinforceable fraction changes sharply with their downstream OPD behavior.
Taken together, the results suggest that student preparation should be evaluated by how it reshapes subsequent learning, not only by how much it improves the student before distillation.
\end{guidance}

\begin{discussion}
\textbf{Guidance.}
For the studied pairing, a brief SFT warm-up is a useful preparation for OPD.
The validation-loss plateau provides a candidate point to check whether additional SFT still improves downstream learning.
\end{discussion}

\subsection{Teacher Adaptation Before Distillation}
\label{subsec:teacher_adaptation}

\begin{figure*}[t!]
\centering
\includegraphics[width=\textwidth]{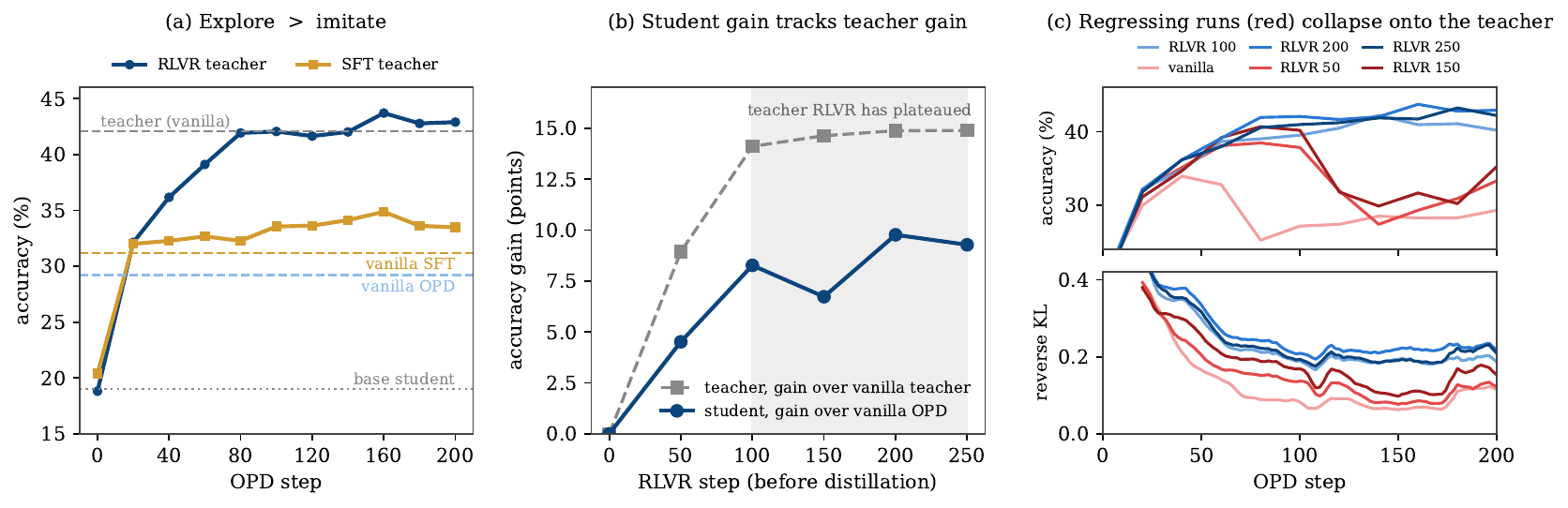}
\vspace{-8mm}
\caption{
\textbf{Teacher adaptation transfers capability, not proximity.}
\textbf{(a)}~OPD from an RLVR-adapted teacher surpasses the off-the-shelf teacher's own accuracy, whereas a SFT-trained teacher improves slightly over vanilla OPD.
\textbf{(b)}~The student's gain tracks the teacher's gain across RLVR checkpoints and flattens once the teacher plateaus (shaded).
\textbf{(c)}~Some runs regress while reverse KL continues to decrease, so teacher matching alone does not explain accuracy retention.
}
\vspace{-5mm}
\label{fig:teacher_adaptation}
\end{figure*}

Student warm-up changes the policy that \emph{receives} the distillation signal and teacher adaptation changes the policy that \emph{produces} it.
We ask: does improving the teacher make it more effective for OPD, and does \emph{how} the teacher is improved matter?
We compare two adaptation methods on the same target-domain prompts: \emph{RLVR}, which improves the teacher on-policy with verifiable outcome rewards, and \emph{STaR-style SFT}~\citep{zelikman2022star}, which trains the teacher to imitate its own verified solutions. \footnote{We also call this SFT teacher, which fine-tuned on one verified-correct self-generated rollout per prompt whenever at least one of four sampled responses is correct.}
Both methods use the same target-domain prompts, and we take checkpoints at $K\in\{50,100,150,200\}$ for both, plus $K=250$ for RLVR.
Each resulting teacher distills the same Qwen3-4B-Base student with identical OPD data, initialization, and budget (Appendix~\ref{app:teacher_adaptation}).

\textbf{Observation 4.4: Teacher capability gains transfer through OPD; a control that does not improve the teacher yields no gain.} \quad 
Distilling from an RLVR-adapted teacher raises student accuracy from $18.9\%$ to $38.5$--$43.7\%$ depending on the checkpoint, against $33.6\%$ from the off-the-shelf teacher (Figure~\ref{fig:teacher_adaptation}a). 
The STaR-adapted teacher does not meaningfully improve ($33.4$--$34.9\%$ across checkpoints), and downstream student accuracy is correspondingly unchanged. 
This confirms that only genuine teacher capability gains transfer.

\textbf{Observation 4.5: The student's gain is a stable fraction of the teacher's gain, so distillation saturates when teacher RLVR does.} \quad 
Teacher accuracy rises from $42.1\%$ before adaptation to $51.0\%$ at $K=50$ and $56.2\%$ at $K=100$, then remains within $56.7$--$57.0\%$.
The corresponding student OPD peaks rise from approximately $33.6\%$ to $38.5\%$ and $42.2\%$, with later checkpoints yielding $40.7$--$43.7\%$ (\Cref{fig:teacher_adaptation}b).
Evaluating DE at these observed peaks, students recover approximately $60\%$ of the gap between the teacher and the initial student.
This complements \S\ref{sec:dynamics}: while scaling teacher size from 14B to 32B did not help, raising the 14B teacher's accuracy from $42\%$ to $57\%$ through RLVR substantially improves transfer.

\begin{guidance}
\textbf{Discussion.}
RLVR and STaR-style self-training see the same prompts but change the teacher differently.
RLVR lets the teacher explore and reinforces what succeeds, raising its accuracy by $9$--$15$ points; self-training imitates solutions the teacher can already produce and leaves its accuracy unchanged.
Only the first gives the student more to learn, in line with evidence that RL generalizes where SFT mainly memorizes~\citep{chu2025sftmemorize} and that OPD transfers what is new to the student~\citep{li2026rethinking}.
The benefit lasts as long as the teacher keeps improving: student gains track teacher gains and flatten once teacher accuracy plateaus.
Our control cannot tell whether RLVR-induced capability would transfer better at matched teacher accuracy.
A stronger teacher also does not prevent regression, since some students lose accuracy while reverse KL keeps falling (\Cref{fig:teacher_adaptation}c).
As a complementary check at fixed teacher size, the checkpoint with the largest $\rho_0^+$ yields the best Qwen3-1.7B-Base student ($29.77\%$ versus $28.00\%$ for minimum rKL), in a single-run comparison where it is also the latest checkpoint (Appendix~\ref{app:teacher_adaptation_diagnostics}).
\end{guidance}

\begin{discussion}
\textbf{Guidance.}
Teacher accuracy is useful to monitor during adaptation, with diminishing downstream returns once it plateaus in this sweep.
Teacher checkpoint selection should also consider the subsequent OPD trajectory, since a high peak need not persist to the end of training.
\end{discussion}

\subsection{Stage Ordering for RLVR Initialization}

\begin{figure*}[h!]
\centering
\vspace{-2mm}
\includegraphics[width=\textwidth]{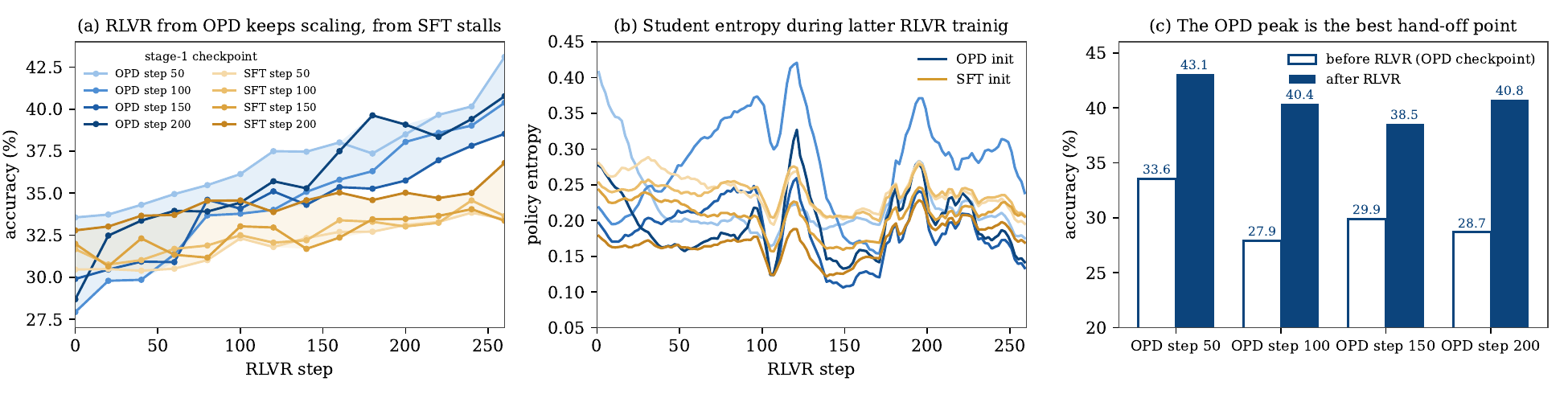}
\vspace{-8mm}
\caption{
\textbf{An OPD checkpoint is a better RLVR initialization than an SFT checkpoint trained on the same trajectories.}
\textbf{(a)}~RLVR from OPD checkpoints (blue) is still rising at step~$260$, while RLVR from SFT checkpoints (gold) stalls by step~$100$. Shading marks the range across checkpoints.
\textbf{(b)}~OPD-initialized runs spend entropy as accuracy rises; SFT-initialized runs keep theirs largely unchanged.
\textbf{(c)}~Among the tested OPD handoffs, the step-$50$ checkpoint has both the highest pre-RLVR accuracy and the highest post-RLVR accuracy, reaching $43.1\%$.
}
\vspace{-2mm}
\label{fig:stage_ordering}
\end{figure*}

\label{subsec:stage_ordering}
We now ask the reverse question: which student state is the better initialization for a subsequent RLVR stage?
To understand that, we compare two separate preparation pipelines, vanilla OPD followed by RLVR and SFT followed by RLVR.
SFT uses verified teacher-generated solutions, whereas OPD uses teacher feedback on student-generated responses.
Both start from Qwen3-4B-Base and use the same fixed Qwen3-14B teacher, followed by the same GRPO procedure on DAPO-Math-17K.
Preparation prompts come from the same retained OpenThoughts-Math pool, with the SFT validation split and other preparation differences detailed in Appendix~\ref{app:stage_ordering_protocol}.
The comparison matters because RLVR is on-policy, so the initialization determines which behaviors are available for exploration and reinforcement, and because current recipes commonly place SFT before RLVR~\citep{yang2025qwen3, xiao2026mimo}.
We repeat it for Qwen3-1.7B-Base and Qwen2.5-1.5B-Base with their corresponding 14B teachers (Appendix~\ref{app:stage_ordering_generalization}).

\textbf{Observation 4.6: OPD initialization enables substantially stronger scaling with RLVR.} \quad
The two initialization families start RLVR at similar accuracy, with SFT slightly ahead on average.
Their trajectories then diverge sharply.
After $260$ RLVR steps, every OPD-initialized run ($38.5$--$43.1\%$) outperforms every SFT-initialized run ($33.4$--$36.8\%$), and the separation grows over the observed horizon: OPD-initialized models are still improving at the end of a full pass over DAPO-Math-17K, while SFT-initialized models gain much less (\Cref{fig:stage_ordering}a).
OPD therefore leaves not merely a stronger checkpoint but \textit{substantially more headroom for scaling the downstream RLVR stage.}

\textbf{Observation 4.7: OPD and SFT checkpoints with comparable entropy yield different RLVR gains.} \quad
A natural explanation is that OPD preserves more policy entropy before RLVR~\citep{aphale2026sft}, but SFT and OPD checkpoints have substantially overlapping initial entropy yet produce clearly separated RLVR trajectories (Figure~\ref{fig:stage_ordering}b). 
Some SFT checkpoints start with higher entropy than OPD checkpoints but still finish lower. 
This agrees with \citet{shen2026stage}, where OPD's higher entropy at RL initialization did not predict a better endpoint, and locates the advantage in something more structured than aggregate uncertainty.

\textbf{Observation 4.8: The highest-accuracy tested OPD checkpoint is also the best RLVR handoff.} \quad
Section~\ref{sec:dynamics} showed that OPD reaches its best task performance well before teacher matching converges.
The same checkpoint also provides the strongest initialization for downstream RLVR.
RLVR from the peak OPD checkpoint (step $50$) reaches $43.11\%$, while later OPD checkpoints that have already regressed finish at $38.5$--$40.8\%$ (Figure~\ref{fig:stage_ordering}c).
RLVR recovers part of the lost performance but does not close the gap.
Continuing OPD beyond its accuracy peak therefore degrades not only the current model but also its potential for subsequent RLVR.

\begin{guidance}
\textbf{Discussion.}
When RLVR follows teacher supervision, our results favor OPD over SFT at comparable student accuracy. One plausible explanation is that OPD provides teacher feedback on the student's own trajectories, aligning the training distribution with the one that RLVR must subsequently explore---but the advantage cannot be reduced to entropy alone. 
Testing whether this advantage grows with model scale, data scale, and substantially larger RLVR compute is a natural next step for frontier post-training systems.
\end{guidance}

\begin{discussion}
\textbf{Guidance.}
When RLVR follows teacher supervision, our results favor OPD over SFT at comparable student accuracy.
OPD should hand off near its task-performance peak rather than continue toward teacher convergence, and scalar entropy alone is not sufficient to choose this transition.
Within the training horizon we study, OPD-initialized models also continue to benefit more from additional RLVR training.
They also reinforce the central composition principle: evaluate a stage by both the capability it adds and the learning conditions it leaves for the next stage.\looseness-1
\end{discussion}

\section{A Simple Composition Recipe}
\label{sec:composition_rule}

\begin{wrapfigure}{r}{0.3\textwidth}
\centering
\vspace{-5mm}
\includegraphics[width=\linewidth]{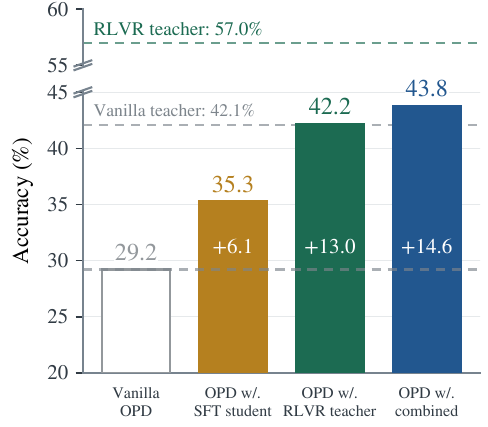}
\vspace{-8mm}
\caption{Final OPD accuracy combined with teacher RLVR adaptation and SFT warm-up.}
\label{fig:composition}
\vspace{-6mm}
\end{wrapfigure}

\S\ref{sec:composition} studied student preparation and teacher adaptation separately.
We now combine them: adapt Qwen3-14B with RLVR until convergence of teacher ($K=200$), briefly warm Qwen3-4B-Base with SFT on the adapted teacher's trajectories ($K=75$), and then distill from that teacher.
At the common $200$-step OPD endpoint, vanilla OPD reaches $29.2\%$ average accuracy, student warm-up reaches $35.3\%$, teacher adaptation reaches $42.2\%$, and both combined reach $43.8\%$ (\Cref{fig:composition}).
The combined pipeline improves over vanilla OPD by $14.6$ percentage points, or $50\%$ relative, at the same OPD step budget with additional preparatory training.
Teacher adaptation provides most of the improvement, while adding student warm-up yields a further $1.6$ points.
The two interventions therefore combine effectively, although their individual gains are not additive.

Together with \S\ref{subsec:stage_ordering}, these results suggest a simple composition principle: \textbf{adapt the teacher with RLVR, briefly warm the student with SFT, distill with OPD, and finally distill to the cold-start student.} More broadly, post-training stages are most effective when each prepares a productive learning interface for the one that follows. Our aim in this work is not a new algorithm but a first controlled map of how these stages interact. Whether the rule holds across model scales, domains, and longer training horizons is beyond our scope, and we hope it invites that study. For interested readers, Appendix~\ref{app:iteration} examines how teacher ordering shapes continued improvement when post-training stages are composed recursively.

\section{Conclusion}
\label{sec:conclusion}

In this work, we presented a controlled investigation into how SFT, RLVR, and OPD interact in LLM post-training.
By changing one stage at a time under a shared protocol, we clarified why the same stage can help or hurt depending on the state it inherits.
Brief SFT improves distillability whereas an RLVR-strengthened student can regress, and teacher RLVR improves transfer until teacher accuracy plateaus.
Combining teacher adaptation with student warm-up raises final OPD accuracy from $29.2\%$ to $43.8\%$ at the same OPD step budget.
Notably, we also show that OPD supports stronger downstream RLVR scaling than SFT within the studied budgets, despite similar starting accuracy and overlapping aggregate entropy in the main Qwen3-4B-Base comparison.
Together, these results show that effective post-training composition depends on both the capability each stage adds and the \emph{learning interface} it leaves for the next stage.

\textbf{Limitations and Future Work.~} Our experiments show that each stage shapes both current capability and subsequent learning, with composition studied mainly using Qwen3-4B and Qwen3-14B. 
Teacher RLVR improves downstream OPD under our protocol, while transferability at matched teacher accuracy and the underlying mechanisms remain open. 
These leave our central finding intact: across controlled interventions, a post-training stage's effect depends on the learning state it inherits.
Future work can extend these findings to larger model scales with mixture-of-expert (MoE) architectures and also agentic learning, where tool use and long-horizon interaction introduce new demands on exploration and transfer.
Our recursive OPD case study (Appendix~\ref{app:iteration}) motivates studying adaptive stage transitions and progressively refreshed teachers as ways to sustain learning as students and tasks evolve.
The broader goal is to develop post-training pipelines that improve current performance while preserving the capacity for continued learning.
We hope this study provides an empirical foundation for post-training pipelines that improve current performance while preparing models for continued learning.\looseness-1

\newpage
\subsection*{AI use statement}

In this work, we used generative AI tools to edit and tighten the manuscript text, to check the consistency of numbers between the text, tables, and figures, to suggest related work incase we missed anything, and to help write and debug plotting scripts.
We did not use generative AI tools to design the experiments, implement the training methods, generate or clean data, or produce or interpret the reported results.
All AI-assisted text and code were reviewed and verified by the authors.
We take responsibility for the final content of this work, including all text, mathematical statements, experimental claims, figures, and references produced with AI assistance.




\subsection*{Ethics statement}


This work studies post-training using open-weight language models, existing reasoning datasets, and model-generated trajectories. 
Our goal is to make interactions between training stages more transparent and accessible to the research community. Distillation may also transfer errors or biases from teachers, so improved benchmark accuracy should not be interpreted as evidence of safe deployment. 
Applications beyond the studied settings require appropriate safety and reliability evaluation.\looseness-1

\subsection*{Reproducibility statement}
We document the experimental procedures and supporting analyses in Appendices~\ref{app:background}--\ref{app:discussion}
Appendix~\ref{app:background} defines the objectives and diagnostics, while Appendix~\ref{app:dynamics} provides model and dataset details, training and evaluation settings, hardware configurations, and metric aggregation.
Appendices~\ref{app:student_warmup}--\ref{app:stage_ordering} describe student warm-up, teacher adaptation, and downstream RLVR initialization, respectively, including data preparation, hyperparameters, and checkpoint selection.
Appendix~\ref{app:iteration}  details the recursive case study, and Appendix~\ref{app:discussion} discusses limitations and unsuccessful attempts.
The main Qwen3-4B-Base scaling experiments use three rollout seeds ($42$, $43$, and $44$) per teacher, with results reported as means and sample standard deviations.
The smaller-student scaling experiments use one run per teacher, and checkpoint sweeps are distinguished from seed replication.
To support independent reproduction and further research, we will release training and evaluation code, data-preparation scripts, experiment configurations, available training logs, evaluation outputs, and scripts for reproducing the tables and figures.\looseness-1




\bibliography{iclr2027_conference}
\bibliographystyle{iclr2027_conference}

\newpage
\appendix
\section*{Appendix}
\addcontentsline{toc}{section}{Appendix} 
\startcontents
\printcontents{}{1}{\setcounter{tocdepth}{2}}
\newpage



\section{Background and Related Work}
\label{app:background}

We first define the three post-training objectives studied throughout the paper (\S\ref{app:objectives}) and the quantities we use to characterize student--teacher transfer (\S\ref{app:metrics}), placing each definition next to the prior work it draws on. We also summarize what is established about OPD and identify the gap this paper studies (\S\ref{app:gap}).

\subsection{Post-Training Objectives}
\label{app:objectives}

\textbf{Supervised fine-tuning (SFT).}
Given a fixed dataset $\mathcal{D} = \{(x^{(i)}, y^{(i)})\}$ of trajectories to imitate, SFT minimizes the token-level cross entropy
\begin{equation}\textstyle
\mathcal{L}_{\text{SFT}}(\theta) = -\mathbb{E}_{(x,y)\sim\mathcal{D}}\left[\sum_{t=1}^{|y|} \log \pi_{S_\theta}(y_t \mid h_t)\right],
\end{equation}
where $h_t = (x, y_{<t})$.
The supervision is off-policy: targets are fixed and do not depend on what the student produces, so the student is trained to reproduce trajectories it may never generate on its own.

\textbf{Reinforcement learning with verifiable rewards (RLVR).}
RLVR optimizes the student on its own rollouts using a scalar outcome reward $r(x,y) \in \{0,1\}$ obtained by verifying the final answer.
We use GRPO~\citep{shao2024deepseekmath}, which samples a group of $G$ responses per prompt and normalizes within the group,
\begin{equation}\textstyle
\hat{A}^{\text{RLVR}}_i = \frac{r(x, y^{(i)}) - \mu_{\text{group}}}{\sigma_{\text{group}}},
\end{equation}
assigning the same advantage to every token of response $i$.
The supervision is on-policy but sparse: one scalar per trajectory, indicating whether the answer was correct but not which tokens were responsible.

\textbf{On-policy distillation (OPD).}
OPD also trains on student-generated trajectories, but replaces the sparse outcome signal with teacher supervision at every generated prefix~\citep{gu2023minillm,agarwal2024policy,lu2025onpolicydistillation}.
Writing $p_t(v) = \pi_S(v \mid h_t)$ and $q_t(v) = \pi_T(v \mid h_t)$, the reverse-KL objective is
\begin{equation}\textstyle
\mathcal{L}_{\text{OPD}} = \mathbb{E}_{x,\, y\sim\pi_S}\left[\frac{1}{|y|}\sum_{t=1}^{|y|} D_{\text{KL}}(p_t \,\|\, q_t)\right].
\end{equation}
Following standard practice~\citep{lu2025onpolicydistillation}, our default implementation uses a sampled-token policy-gradient estimator, which requires only one teacher log-probability per sampled token rather than the full teacher distribution.
For the token $y_t$ sampled by the student we define the teacher advantage
\begin{equation}
A_t = \texttt{sg}\!\left[\log \pi_T(y_t \mid h_t) - \log \pi_{S_{\text{old}}}(y_t \mid h_t)\right],
\label{eq:adv}
\end{equation}
where $\texttt{sg}[\cdot]$ denotes the stop-gradient operation, and optimize it with the same clipped importance-ratio objective used for RLVR.
The sign of $A_t$ determines the effect of each update.
When $A_t > 0$ the update reinforces a student-sampled token that the teacher prefers more strongly.
When $A_t < 0$ it suppresses that token without specifying which alternative should replace it.
The mean of $A_t$ under the student's own samples equals the negative reverse KL at that prefix (Appendix~\ref{app:dynamics}, Proposition~1), so the mean carries no information beyond the loss itself.
The mean therefore carries no information beyond reverse KL, motivating diagnostics that capture other properties of the distribution of $A_t$, including its sign structure.

Recent work has questioned the bias, variance, and robustness of this sampled-token estimator~\citep{zhao2026opd+,oh2026kl,xing2026trust,fu2026revisiting}.
To test whether our observations depend on this formulation, we additionally evaluate a dense top-$k$ forward-KL objective,
\begin{equation}\textstyle
\mathcal{L}^{(k)}_{\text{FKL}} = \mathbb{E}_{x,\, y\sim\pi_S}\left[\frac{1}{|y|}\sum_{t=1}^{|y|} \sum_{v \in \mathcal{V}^T_k(h_t)} q_t(v)\log\frac{q_t(v)}{p_t(v)}\right], \quad \mathcal{V}^T_k(h_t) = \text{TopK}(q_t, k),
\label{eq:fkl}
\end{equation}
which matches the teacher's distribution over its $k$ most likely tokens and therefore does specify replacements.

The three methods differ along two axes: SFT provides dense supervision on fixed trajectories, RLVR provides sparse outcome supervision on student trajectories, and OPD provides dense teacher supervision on student trajectories.

\subsection{Measuring Transfer}
\label{app:metrics}

We characterize student--teacher transfer from three views: the outcome of a completed distillation run, the structure of the teacher's signal on the student's behavior at initialization, and the on-policy distance between the two policies during training.

\textbf{Distillation effectiveness.}
Let $A_S$ and $A_{S'}$ denote the student's accuracy before and after distillation, and $A_T$ the teacher's accuracy.
We define
\begin{equation}\textstyle
\text{DE}(S \leftarrow T) = \frac{A_{S'} - A_S}{A_T - A_S},
\end{equation}
the fraction of the teacher--student gap recovered through distillation, with $1-\mathrm{DE}$ the \emph{unrecovered share}.
The quantity is well defined when the teacher outperforms the student ($A_T > A_S$), and negative values indicate regression.
The same quantity is used concurrently by \citet{li2026rethinking} under the name \emph{gap recovery rate}.
We adopt it because raw accuracy gains are not comparable across students that begin at different accuracies.

\textbf{Reinforceable fraction.}
Partitioning sampled tokens by the sign of $A_t$ in Eq.~\ref{eq:adv}, we define
\begin{equation}
\rho_0^+ = \mathbb{P}_{y\sim\pi_{S_0}}\left[A_t > 0\right],
\end{equation}
the fraction of tokens on the student's rollout that the teacher would reinforce rather than suppress, evaluated at initialization before any parameter update.
The same sign decomposition is introduced concurrently by \citet{wang2026demystifying}, who use it to diagnose signal corruption and to motivate advantage clipping.
\citet{li2026rethinking} track a closely related quantity, the top-$k$ overlap between student and teacher at student-visited states, and show that it separates successful from failing runs.
Our use differs in purpose rather than in construction.
Because $\rho_0^+$ depends only on the student's initial distribution, it requires a single scoring pass and no training, and therefore gives an initialization-time view of how much of the student's current on-policy behavior the teacher can directly reinforce.
In \S\ref{sec:dynamics} we test whether this quantity, measured before any training, identifies which teacher will distill most effectively.
Pre-training predictors of distillation gain have also been proposed in the self-distillation setting~\citep{he2026predictive}.

\textbf{On-policy teacher--student divergence.}
We also measure the reverse KL on student-generated trajectories after $s$ optimizer steps,
\begin{equation}\textstyle
\widehat{\text{rKL}}_s = \mathbb{E}_{x,\, y\sim\pi_{S_s}}\left[\frac{1}{|y|}\sum_t \log\frac{\pi_{S_s}(y_t \mid h_t)}{\pi_T(y_t \mid h_t)}\right],
\qquad
\kappa_s = \widehat{\text{rKL}}_s / \widehat{\text{rKL}}_0 .
\end{equation}
The first measures teacher--student distance on the states the student actually visits, and provides the natural reference for asking whether distillation effectiveness can be explained by scalar policy divergence alone.
The normalized form $\kappa_s$ places pairs with different initial divergence on a common axis and is what we plot when studying training dynamics.
Both are computed from the student's rollouts on training prompts, so they can be monitored during training without additional evaluation passes.

\subsection{What Is Known, and What Is Not}
\label{app:gap}

\paragraph{OPD Training Dynamics Landscape}
Increasing teacher capability alone does not guarantee more effective distillation.
Classical knowledge-distillation work observed that higher teacher accuracy need not produce a better student~\citep{cho2019efficacy}, motivating intermediate teacher assistants~\citep{mirzadeh2020improved} and revised objectives for strong teachers~\citep{huang2022knowledge}.
The effect reappears for reasoning models, where traces from strong teachers can underperform simpler alternatives for small students~\citep{li2025small}.
In the on-policy regime, three concurrent studies reach converging conclusions.
\citet{li2026rethinking} identify thinking-pattern consistency and genuine novelty of teacher knowledge as two important conditions for successful transfer, and show that successful runs are marked by rising top-$k$ overlap at student-visited states.
\citet{wang2026demystifying} characterize OPD as an exploration catalyst rather than a capability expander and attribute failures to signal corruption under student--teacher mismatch and to length exploitation.
\citet{zhu2026many} document pitfalls of the top-$K$ reverse-KL approximation and of locally incompatible supervision on drifted prefixes.
Section~\ref{sec:dynamics} re-examines these findings under a single controlled protocol across a $3\times3$ grid of student and teacher sizes, and tests robustness to model family, distillation objective, and training data.

\paragraph{Interventions on Individual Training Stages}
Prior work has also examined each stage adjacent to OPD in isolation.
On the student side, an off-policy cold start (SFT on teacher-generated rollouts before distillation) raises initial overlap and improves both early optimization and the final ceiling~\citep{li2026rethinking,zhu2026many}.
On the teacher side, RLVR adaptation of the teacher improves downstream distillation and moves the teacher closer to the student's distribution~\citep{zhu2026many}, rejection fine-tuning has been used to calibrate teachers for distillability~\citep{zhan2026distillation}, and the teacher's RL-induced policy shift can itself be transferred in place of the teacher policy~\citep{feng2026weak}.
On the stage that follows, pre-RL policy entropy has been shown to predict post-RLVR outcomes better than pre-RL accuracy~\citep{aphale2026sft}, over-trained SFT checkpoints can lose plasticity needed for downstream RL~\citep{liu2026rl}, and SFT, OPD, and RLVR leave structurally different updates in parameter space~\citep{shen2026geometry}.
A concurrent study of two-stage post-training for vision-language models compares SFT and OPD warm-starts directly and finds that OPD's higher entropy at RL initialization does not translate into a better endpoint in their setting~\citep{shen2026stage}.
We return to this contrast in \S\ref{sec:composition}.

\paragraph{The Gap We Think}
Prior work has established several effects of individual stages and adjacent stage transitions. 
What remains much less understood is how these stages \emph{compose}, that is, how changing one stage alters the learning conditions faced by the next.
We first characterize the OPD dynamics that determine whether a student--teacher pairing is productive (\S\ref{sec:dynamics}), then study how SFT, RLVR, and OPD reshape that pairing when composed (\S\ref{sec:composition}), and finally ask whether it can be maintained recursively as the student evolves (\S\ref{app:iteration}).


\section{Experimental Details and Additional Analyses on OPD Dynamics}
\label{app:dynamics}

This appendix details \S\ref{sec:dynamics} with complete configurations, learning curves, and additional comparisons. 
The analyses examine how teacher scale interacts with the student state, distillation objective, and training duration. 
For interested readers, we also provide more context on the reinforceable fraction and the conditions under which an early handoff can preserve task performance.

\subsection{Experimental Setup and Training Configurations} 
\label{app:dynamics:setup} 

Our main training grid pairs Qwen3-{0.6B, 1.7B, 4B}-Base students with Qwen3-{8B, 14B, 32B} post-trained teachers, systematically varying both model sizes to characterize how OPD dynamics depend on student and teacher scale.
All nine training runs use DAPO-Math-17k-Processed\footnote{\url{https://huggingface.co/datasets/open-r1/DAPO-Math-17k-Processed}} with the sampled-token reverse-KL objective during distillation~\citep{lu2025onpolicydistillation}, built on verl as our base repository~\citep{sheng2024hybridflow-verl}.
Table~\ref{tab:dynamics_opd_config} lists the shared training configuration used across all scaling experiments.
For our main experiments with the largest student, Qwen3-4B-Base, we report three runs with seeds 42, 43, and 44 (as shown in \Cref{fig:ceiling}) to ensure experimental reliability.
Qwen3-0.6B-Base and Qwen3-1.7B-Base  comparisons use a single run per teacher due to computational cost. 
Since the 4B experiments already establish that our main claim is robust to seed variation, these smaller scales serve primarily to test whether the observation generalizes across student sizes.

The main manuscript uses step 200 as the primary comparison point to reflect our computational budget. 
However, during our internal experiments, some runs continued beyond this point, and we prefer to include their longer trajectories here to provide a fuller picture of the training dynamics for transparency, so that interested readers can build on them in future work.
That said, our core runs follow a one-epoch schedule with an optimizer horizon of 279 steps. We report results at step 260, their last shared evaluation, alongside step 200 for comparison with the main text. 
The forward-KL experiments are compared at step 180, which is available for every teacher and every value of $k$; where as model-state and training-data comparisons use step 200. 
Each table specifies its evaluation horizon, and we report variability across runs only where multiple seeds are available.

\begin{table}[htbp]
\centering
\small
\caption{
Shared OPD configuration for the main experiment grid in \S\ref{sec:dynamics}.}
\label{tab:dynamics_opd_config}
\begin{tabular}{@{}lp{0.5\linewidth}@{}}
\toprule
\textbf{Setting} & \textbf{Value} \\ \midrule
Training data & DAPO-Math-17k \\
Parameter update & Full-parameter training, LoRA rank $0$ \\
Optimizer & AdamW \\
Learning rate / $(\beta_1,\beta_2)$ & $10^{-6}$ / $(0.9,0.999)$ \\
Schedule / warm-up & Constant learning rate / 10 steps \\
Weight decay / gradient norm clipping & $0.01$ / $1.0$ \\
Prompt batch / PPO minibatch & 64 / 64 \\
PPO epochs / clipping ratio & 1 / $0.2$ \\
Rollout responses / temperature / top-$p$ & 4 / $1.0$ / $1.0$ \\
Maximum prompt / response length & 2048 / 4096 tokens \\
Loss mode / policy-gradient update & Sampled-token reverse KL (\texttt{k1}) / enabled \\
Distillation coefficient / task reward & $1.0$ / disabled \\
Auxiliary KL reward / KL loss & Disabled / disabled \\
Loss aggregation / entropy coefficient & Token mean / $0$ \\
Configured loss ceiling / log-probability floor & $10$ / $-10$ \\
Data shuffling / overlong-prompt filtering & Disabled / enabled \\
Qwen3 thinking mode & Disabled \\
Actor backend / rollout backend & FSDP / vLLM \\
Actor / teacher GPU allocation & 4 / 4 GPUs \\
Rollout / teacher tensor parallelism & 1 / 1 \\
Sequence parallelism / compute dtype & 1 / bfloat16 \\
Dynamic batching / actor token budget & Enabled / 24,576 tokens per GPU \\
Teacher inference token budget & 4096 tokens per batch \\
Training schedule & One epoch with no explicit trainer step cap \\
Evaluation interval & Every 20 steps \\
\bottomrule
\end{tabular}
\end{table}

\begin{table}[htbp]
\centering\small
\caption{Forward-KL configuration with optimization, sampling, and length settings match Table~\ref{tab:dynamics_opd_config}. The objective uses teacher top-$k$ support.}
\label{tab:dynamics_fkl_config}
\begin{tabular}{@{}lp{0.55\linewidth}@{}}
\toprule
\textbf{Setting} & \textbf{Value} \\
\midrule
Student / teachers & Qwen3-4B-Base / Qwen3-\{8B, 14B, 32B\} \\
Training data & DAPO-Math-17k \\
Loss mode / top-$k$ & \texttt{forward\_kl\_topk} / $k\in\{16,32,64\}$ \\
Update & Direct distillation loss with policy-gradient mode disabled \\
Optimizer / learning rate & AdamW / $10^{-6}$ \\
Schedule / warm-up & Constant learning rate / 10 steps \\
Prompt batch / responses per prompt & 64 / 4 \\
Maximum prompt / response length & 2,048 / 4096 tokens \\
\bottomrule
\end{tabular}
\end{table}

\begin{table}[htbp]
\centering
\small
\caption{Evaluation protocol where the primary aggregate assigns equal weight to each of the seven benchmarks.}
\label{tab:dynamics_eval_config}
\begin{tabular}{@{}lp{0.55\linewidth}@{}}
\toprule
\textbf{Setting} & \textbf{Value} \\
\midrule
Primary benchmarks & AIME 2024, AIME 2025, AMC 2023, MATH-500, Minerva Math, OlympiadBench, GPQA-Diamond \\
Responses per question & 4 \\
Temperature / top-$p$ / top-$k$ & $0.6$ / $0.95$ / disabled \\
Prompt / response limit & 2,048 / 32,768 tokens \\
Per-benchmark score & Mean correctness over four responses (\texttt{mean@4}) \\
Configured evaluator & \texttt{ttrl\_math.reward\_func} in \texttt{verl} \\
Primary aggregate & Unweighted mean of the seven benchmark scores \\
Initial accuracy & Evaluation at training step 0 \\
Initial training diagnostics & First logged training step, recorded as step 1 \\
Across-run summary & Mean and sample standard deviation where $n=3$ \\
\bottomrule
\end{tabular}
\end{table}

\begin{table}[t]
\centering
\small
\caption{Training and evaluation datasets, prompt lengths use the Qwen3 tokenizer with the training chat template.}
\label{tab:data_stats}
\begin{tabular}{lrr}
\toprule
\textbf{Dataset} & \textbf{Unique prompts} & \textbf{Average prompt tokens} \\
\midrule
\multicolumn{3}{l}{\textit{Training}} \\
DAPO-Math-17k & 17,917 & 122 \\
OpenThoughts3-Math & 30,000 & 113 \\
\midrule
\multicolumn{3}{l}{\textit{Evaluation}} \\
AIME 2024 & 30 & 150 \\
AIME 2025 & 30 & 145 \\
AMC 2023 & 83 & 119 \\
MATH-500 & 500 & 98 \\
Minerva Math & 272 & 158 \\
OlympiadBench & 675 & 127 \\
GPQA-Diamond & 198 & 240 \\
\bottomrule
\end{tabular}
\end{table}

\subsection{Metric Definitions and Aggregation}
\label{app:dynamics:metrics}

For benchmark $b$ with $N_b$ questions, let $c_{bqj}$ indicate whether sampled response $j$ to question $q$ is correct. 
The benchmark score and primary aggregate are:
\begin{equation} 
A_b=\frac{100}{4N_b}\sum_{q=1}^{N_b}\sum_{j=1}^{4}c_{bqj}, \qquad A=\frac{1}{7}\sum_{b=1}^{7}A_b
\label{eq:app_b_accuracy} 
\end{equation} 
Thus, \texttt{mean@4} measures average correctness across sampled responses. 
It does not apply majority voting or count a question as solved whenever one of its four responses is correct.

Let $A_{S\leftarrow T,h}$ denote the student's accuracy after $h$ OPD steps. 
We compute gain and distillation efficiency using: 
\begin{equation} 
G_h=A_{S\leftarrow T,h}-A_{S,0}, \qquad \mathrm{DE}_h=\frac{A_{S\leftarrow T,h}-A_{S,0}}{A_T-A_{S,0}} 
\label{eq:app_b_de} 
\end{equation} 
The initial student score is pooled across the runs in the corresponding main-grid row. 
These baselines are $2.25\%$, $10.79\%$, and $18.89\%$ for the 0.6B, 1.7B, and 4B students. 
The fixed teacher reference accuracies used for DE are $40.09\%$, $42.11\%$, and $42.80\%$ for the 8B, 14B, and 32B teachers. 
We report absolute accuracy alongside DE because a larger teacher reference increases the denominator of \Cref{eq:app_b_de}. 
The two quantities therefore answer different questions about transfer.

For a student-generated token $y_t$ at prefix $h_t$, define the log-probability advantage as
\begin{equation}
a_t=\log p_T(y_t\mid h_t)-\log p_S(y_t\mid h_t)
\end{equation}
Within a valid response-token block $\mathcal B$, the reinforceable fraction is
\begin{equation}
\widehat\rho^+_{\mathcal B} =\frac{\sum_{t\in\mathcal B}m_t\mathbf{1}[a_t>0]} {\sum_{t\in\mathcal B}m_t}
\label{eq:app_b_rho}
\end{equation}
where $m_t$ is the response mask.
The reported statistic follows the trainer's mean aggregation of these masked token statistics, while zero advantages are excluded from both the positive and negative fractions.
The term \emph{reinforceable} refers to the sign of the distillation signal and it does not classify the correctness of a solution.
On the other hand, rKL curves use the sampled-token statistic logged by the same training code.

All accuracy summaries first aggregate benchmarks within a run and then aggregate runs.
Error bars represent sample standard deviation across rollout seeds.
For peak summaries, we take the maximum within each run before averaging.
These observed maxima provide retrospective reference values and they are distinct from selecting a checkpoint using an independent validation set.

\subsection{Full Student--Teacher Scaling Results}
\label{app:dynamics:scaling}

\begin{wraptable}{r}{0.52\textwidth}
\centering
\footnotesize
\caption{Complete Qwen3 base-student grid where accuracy is the seven-benchmark mean in percent. Gain and DE use the pooled initial score for each student.}
\label{tab:dynamics_grid}
\setlength{\tabcolsep}{3pt}
\begin{tabular}{llrrrr}
\toprule
Student & Teacher & $A_{200}$ & $A_{260}$ & Gain$_{260}$ & DE$_{260}$ (\%) \\
\midrule
0.6B & 8B  & 14.29 & 14.91 & +12.65 & 33.4 \\
0.6B & 14B & 15.00 & 13.24 & +10.99 & 27.6 \\
0.6B & 32B  & 13.66 & 13.24 & +10.99 & 27.1 \\
\midrule
1.7B & 8B  & 23.00 & 22.26 & +11.47 & 39.1 \\
1.7B & 14B  & 20.95 & 20.13 & +9.34 & 29.8 \\
1.7B & 32B  & 20.34 & 21.36 & +10.58 & 33.0 \\
\midrule
4B & 8B  & 32.70 & 31.49 & +12.60 & 59.4 \\
4B & 14B & 29.23 & 29.19 & +10.30 & 44.4 \\
4B & 32B  & 29.25 & 28.57 & +9.68 & 40.5 \\
\bottomrule
\end{tabular}
\vspace{-5mm}
\end{wraptable}
Table~\ref{tab:dynamics_grid} reports the complete grid at two training horizons. 
At step 260, the 8B teacher produces the highest accuracy for all three students. 
For Qwen3-4B-Base, its $31.49\%$ accuracy exceeds the 14B and 32B outcomes by $2.30$ and $2.92$ points. 
The corresponding DE values show that increased teacher scale does not consistently translate into greater recovery of the teacher's capability.

\begin{figure}[htbp]
\centering
\includegraphics[width=\linewidth]{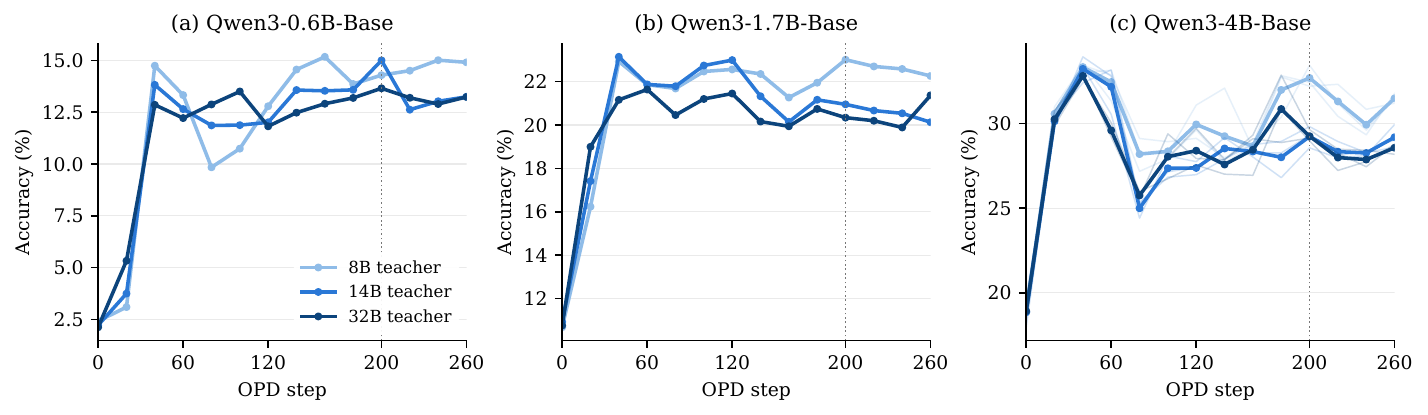}
\caption{\textbf{Complete OPD learning curves across student and teacher sizes.}
Each panel compares three teachers for one base student.
The 0.6B and 1.7B curves show individual runs.
For 4B, bold lines show the three-run mean and faint lines show individual runs.
The dotted line marks step 200.
All curves extend to the shared final evaluation at step 260.}
\label{fig:app_b_learning_curves}
\end{figure}

Figure~\ref{fig:app_b_learning_curves} shows that this endpoint comparison
coexists with different learning dynamics.
The 0.6B student undergoes a pronounced intermediate decline with the 8B
teacher before recovering.
At step 200, the 14B teacher instead leads this row by $0.71$ points.
The 1.7B and 4B students favor the 8B teacher at both reported horizons.
For 4B, a strong early improvement is followed by a decline and partial recovery.
These results suggest that teacher selection should be interpreted relative
to the intended training budget.
They also show why a single endpoint does not fully describe transfer.

\subsection{Robustness Across Distillation Objectives}
\label{app:dynamics:objectives}

\begin{wraptable}{r}{0.5\textwidth}
\centering
\footnotesize
\vspace{-3mm}
\caption{Forward-KL results. All cells use Qwen3-4B-Base and one run and peak accuracy is restricted to the same budget. Baseline and teacher references match \Cref{tab:dynamics_grid}.}
\label{tab:dynamics_fkl_results}
\vspace{2pt}
\setlength{\tabcolsep}{3pt}
\begin{tabular}{rlrrrr}
\toprule
$k$ & Teacher & $A$ (\%) & Gain & DE (\%) & Peak (\%) \\
\midrule
16 & 8B & 33.66 & +14.77 & 69.7 & 34.43 \\
16 & 14B & 34.48 & +15.59 & 67.2 & 34.48 \\
16 & 32B & 31.79 & +12.91 & 54.0 & 33.48 \\
\midrule
32 & 8B & 34.22 & +15.33 & 72.3 & 35.11 \\
32 & 14B & 34.39 & +15.50 & 66.8 & 35.28 \\
32 & 32B & 32.64 & +13.75 & 57.5 & 33.12 \\
\midrule
64 & 8B & 35.09 & +16.20 & 76.4 & 35.09 \\
64 & 14B & 34.44 & +15.55 & 67.0 & 35.56 \\
64 & 32B & 32.98 & +14.09 & 58.9 & 33.68 \\
\bottomrule
\end{tabular}
\end{wraptable}
To assess whether our teacher-scaling observations extend across distillation objectives, we further examine Qwen3-4B-Base under forward KL with teacher top-$k$ support for $k\in{16,32,64}$. \Cref{tab:dynamics_fkl_results} compares all nine configurations at a common budget of 180 steps, showing that distillation from Qwen3-32B yields the lowest accuracy and DE at every value of $k$. Among smaller teachers, Qwen3-14B achieves higher accuracy at $k=16$ and $k=32$, while Qwen3-8B performs better at $k=64$. 
These results extend our finding that larger teachers do not consistently improve transfer, while showing that the preferred teacher can depend on how much of its output distribution is matched.

This comparison also highlights why accuracy and DE should be considered together. 
At $k=16$, distillation from Qwen3-14B yields a student accuracy of $34.48\%$ at step 180, compared with $33.66\%$ when using Qwen3-8B. 
Despite this accuracy advantage, Qwen3-14B has lower DE because its larger capability gap over the initial student increases the normalization denominator. 
A larger teacher can therefore produce a more accurate student while transferring a smaller fraction of its capability advantage, indicating that teacher selection depends on both the distillation objective and how transfer is measured.

\subsection{Model States, Training Data, and Evaluation Domains}
\label{app:dynamics:generalization}

\paragraph{Student state and model family.}
To examine how OPD effectiveness varies across model families and student starting states, \Cref{tab:dynamics_student_states} extends our analysis to post-trained Qwen3 students and both base and instruct Qwen2.5 students at a common budget of 200 steps. 
For Qwen2.5-1.5B, OPD improves base-student accuracy by $18.54$ to $19.14$ points across three teachers, compared with gains of $2.48$ to $2.66$ points for its instruct counterpart. 
In contrast, post-trained Qwen3-4B starts at $39.58\%$ accuracy and finishes below this baseline with every teacher. 
These results show that OPD gains depend substantially on the student's starting state, extending this observation beyond our main Qwen3 base-model grid.

\begin{table}[htbp]
\centering
\small
\caption{
Student-state and model-family comparisons after 200 OPD steps. 
Entries report accuracy (\%) with gains over the pooled initial student score in parentheses. 
The smallest teacher is 8B for Qwen3 and 7B for Qwen2.5. 
Results average three runs per teacher for Qwen3-4B-Base and use one run per configuration otherwise.
}
\label{tab:dynamics_student_states}
\setlength{\tabcolsep}{4pt}
\begin{tabular}{llrccc}
\toprule
\textbf{Student} & \textbf{State} & \textbf{Initial} & \textbf{Qwen2.5-7B/Qwen3-8B teacher} & \textbf{14B teacher} & \textbf{32B teacher} \\
\midrule
Qwen3-1.7B & Base & 10.79 & 23.00 (+12.22) & 20.95 (+10.16) & 20.34 (+9.55) \\
Qwen3-1.7B & Post-trained & 29.51 & 29.82 (+0.31) & 31.17 (+1.65) & 29.18 (-0.33) \\
Qwen3-4B & Base & 18.89 & 32.70 (+13.81) & 29.23 (+10.34) & 29.25 (+10.36) \\
Qwen3-4B & Post-trained & 39.58 & 38.03 (-1.55) & 39.28 (-0.31) & 37.69 (-1.89) \\
Qwen2.5-1.5B & Base & 1.30 & 19.85 (+18.54) & 20.01 (+18.71) & 20.44 (+19.14) \\
Qwen2.5-1.5B & Instruct & 18.08 & 20.69 (+2.61) & 20.56 (+2.48) & 20.73 (+2.66) \\
Qwen2.5-3B & Base & 18.61 & 25.87 (+7.27) & 26.10 (+7.50) & 25.28 (+6.68) \\
Qwen2.5-3B & Instruct & 25.11 & 24.40 (-0.70) & 25.24 (+0.14) & 25.56 (+0.45) \\
\bottomrule
\end{tabular}
\end{table}

A more capable post-trained student has less measured headroom for improvement, which can change how much it benefits from teacher supervision and motivates our study of student preparation in \S\ref{sec:composition}. 
Teacher rankings also vary across model families, as illustrated by Qwen2.5-1.5B-Base, which achieves its highest step-200 accuracy with the 32B teacher. 
These results suggest that teacher selection should account for the student's model family and starting state rather than relying on a fixed ranking of teachers.

\paragraph{Teacher training state.}
\Cref{tab:dynamics_teacher_states} compares base and post-trained Qwen3 teachers at matched model sizes, with post-trained teachers yielding higher peak student accuracy in four of five pairs. 
For Qwen3-4B-Base, peak accuracy increases from $31.52\%$ to $33.38\%$ with an 8B teacher and from $32.35\%$ to $33.49\%$ with a 14B teacher. 
The exception is Qwen3-1.7B-Base paired with a 14B teacher, where the base teacher yields a higher peak. 
Endpoint rankings also depend on how much of this improvement persists through training, suggesting that teacher training state can affect both peak performance and its retention. 
This distinction motivates our controlled teacher-adaptation experiments in Section~\ref{sec:composition}.

\begin{table}[htbp]
\centering
\small
\caption{
Comparison of base and post-trained Qwen3 teachers at matched model sizes, using base students throughout.
Endpoint accuracy is measured at step 200, and peaks are computed over evaluations through that step. 
Results with post-trained teachers average three runs for the 4B student, while all other configurations use one run.
}
\label{tab:dynamics_teacher_states}
\setlength{\tabcolsep}{4pt}
\begin{tabular}{llrrrr}
\toprule
\textbf{Student} & \textbf{Teacher size} & \textbf{Base $A$} & \textbf{Post $A$} & \textbf{Base peak} & \textbf{Post peak} \\
\midrule
1.7B & 8B & 19.66 & 23.00 & 21.50 & 23.00 \\
1.7B & 14B & 22.30 & 20.95 & 23.54 & 23.14 \\
4B & 8B & 25.05 & 32.70 & 31.52 & 33.38 \\
4B & 14B & 30.81 & 29.23 & 32.35 & 33.49 \\
8B & 14B & 31.19 & 29.34 & 33.32 & 35.10 \\
\bottomrule
\end{tabular}
\end{table}

\begin{wraptable}{r}{0.52\textwidth}
\centering
\small
\caption{Qwen3-4B-Base accuracy (\%) by distillation dataset.
DAPO-Math reports mean $\pm$ s.d.\ over three runs.
OpenThoughts-Math is run per teacher.}
\label{tab:dynamics_training_data}
\setlength{\tabcolsep}{5pt}
\renewcommand{\arraystretch}{1.1}
\vspace{-0.5mm}
\begin{tabular*}{\linewidth}{@{\extracolsep{\fill}}lccc@{}}
\toprule
& DAPO-Math & \multicolumn{2}{c}{OpenThoughts-Math} \\
\cmidrule(lr){2-2}\cmidrule(l){3-4}
Teacher & Step 200 & Step 200 & Step 240 \\
\midrule
8B  & $32.70 \pm 0.66$ & 31.29 & 31.10 \\
14B & $29.23 \pm 0.63$ & 28.80 & 29.18 \\
32B & $29.25 \pm 0.34$ & 28.00 & 30.35 \\
\bottomrule
\end{tabular*}
\vspace{-1mm}
\end{wraptable}
\paragraph{Training distribution.}
To examine whether our teacher-scaling observations depend on the distillation data, Table~\ref{tab:dynamics_training_data} compares Qwen3-4B-Base trained on DAPO-Math and OpenThoughts-Math at a common budget of 200 OPD steps.
Qwen3-8B yields the highest student accuracy on both datasets, reaching $32.70\%$ on DAPO-Math and $31.29\%$ on OpenThoughts-Math. 
This agreement indicates that its advantage in this student setting extends beyond a single training distribution. 
At step 240 on OpenThoughts-Math, Qwen3-8B remains the best teacher, although Qwen3-32B overtakes Qwen3-14B. 
The comparison therefore supports a consistent preference for the 8B teacher across these two datasets, while showing that the relative performance of larger teachers can change with training duration.

\paragraph{Evaluation domain.}
To examine whether teacher rankings depend on the evaluation domain, Table~\ref{tab:dynamics_benchmarks} reports benchmark-level results for Qwen3-4B-Base at step 260. Qwen3-8B achieves the highest mean accuracy on all six math benchmarks, leading both the math-six aggregate at $33.22\%$ and our primary core-seven aggregate at $31.49\%$. 
Qwen3-14B instead performs best on GPQA-Diamond and achieves the highest ten-benchmark average at $28.14\%$ after including SciKnowEval. 
These results show that the preferred teacher depends on the target task mixture, even for a fixed student and training recipe.
We retain core-seven as the primary metric and use the additional benchmarks to identify where teacher rankings differ across domains.

\begin{table}[htbp]
\centering
\small
\caption{Qwen3-4B-Base accuracy (\%) at step 260, reported as
mean $\pm$ sample standard deviation over three rollout seeds.
Aggregates average the first six, first seven, or all ten benchmarks
with equal weight.}
\label{tab:dynamics_benchmarks}
\setlength{\tabcolsep}{6pt}
\renewcommand{\arraystretch}{1.1}
\begin{tabular}{lrrr}
\toprule
Benchmark / aggregate & 8B teacher & 14B teacher & 32B teacher \\
\midrule
AIME 2024 & $13.89\pm1.73$ & $13.61\pm2.68$ & $13.33\pm0.83$ \\
AIME 2025 & $12.78\pm0.48$ & $12.50\pm0.83$ & $11.11\pm1.27$ \\
AMC 2023 & $42.97\pm1.06$ & $39.86\pm0.92$ & $36.45\pm1.59$ \\
MATH-500 & $72.23\pm0.70$ & $61.57\pm0.70$ & $64.13\pm1.25$ \\
Minerva Math & $21.57\pm0.70$ & $18.29\pm0.09$ & $17.68\pm0.41$ \\
OlympiadBench & $35.89\pm0.94$ & $30.83\pm0.97$ & $31.00\pm0.13$ \\
GPQA-Diamond & $21.11\pm0.51$ & $27.71\pm0.86$ & $26.31\pm1.68$ \\
SciKnowEval-Biology & $14.67\pm5.20$ & $15.17\pm1.04$ & $15.33\pm1.26$ \\
SciKnowEval-Chemistry & $18.25\pm2.00$ & $23.75\pm0.25$ & $20.00\pm1.09$ \\
SciKnowEval-Physics & $16.04\pm0.18$ & $38.12\pm1.13$ & $32.40\pm6.59$ \\
\midrule
Math-six & $33.22\pm0.39$ & $29.44\pm0.87$ & $28.95\pm0.63$ \\
Core-seven & $31.49\pm0.29$ & $29.19\pm0.68$ & $28.57\pm0.34$ \\
All ten & $26.94\pm0.51$ & $28.14\pm0.44$ & $26.77\pm0.38$ \\
\bottomrule
\end{tabular}
\end{table}

\subsection{Reinforceable Fraction and Teacher Selection}
\label{app:dynamics:selection}

\begin{table}[htbp]
\centering
\small
\caption{Initial reverse KL and $\rho_0^+$ for each student--teacher pair.}
\label{tab:dynamics_initial_stats}
\begin{tabular*}{\linewidth}{@{\extracolsep{\fill}}lrrrrrrrrr@{}}
\toprule
Student & \multicolumn{3}{c}{0.6B} & \multicolumn{3}{c}{1.7B} & \multicolumn{3}{c@{}}{4B} \\
\cmidrule(lr){2-4} \cmidrule(lr){5-7} \cmidrule(l){8-10}
Teacher & 8B & 14B & 32B & 8B & 14B & 32B & 8B & 14B & 32B \\
\midrule
Initial rKL & 0.740 & 0.680 & 0.648 & 0.447 & 0.478 & 0.426 & 0.370 & 0.376 & 0.382 \\
$\rho_0^+$ (\%) & 37.32 & 33.20 & 29.39 & 53.34 & 49.12 & 34.37 & 54.79 & 47.83 & 33.11 \\
\bottomrule
\end{tabular*}
\end{table}

To relate teacher selection to the initial student--teacher pairing, \Cref{tab:dynamics_initial_stats} reports reinforceable fraction and reverse KL from the same runs used in our accuracy comparisons. 
The reinforceable fraction decreases with teacher size for every student, falling from $54.79\%$ to $33.11\%$ for Qwen3-4B-Base as teacher size increases from 8B to 32B. 
Reverse KL follows a different pattern: for both 0.6B and 1.7B students, Qwen3-32B has the lowest initial rKL but does not yield the highest accuracy at either step 200 or step 260. 
A teacher's proximity to the student therefore does not necessarily identify the most effective source of supervision, motivating a direct comparison of selection rules.

We quantify each teacher choice through selection regret, defined as the difference between student accuracy obtained with the best observed teacher and that obtained with the selected teacher:
\begin{equation}
\mathcal R_h(S,\widehat T)
=
\max_{T\in\mathcal T} A_{S\leftarrow T,h}
-
A_{S\leftarrow\widehat T,h},
\end{equation}
where $\mathcal T$ contains the three candidate teachers, $h$ denotes the number of OPD steps, and $A_{S\leftarrow T,h}$ is the resulting student accuracy in percent. 
For the replicated 4B experiments, outcomes and initial diagnostics are averaged across runs before applying each selection rule.

\Cref{tab:dynamics_selection} compares four rules at steps 200 and 260. 
Selecting the largest initial reinforceable fraction identifies the best-performing teacher for all three students at step 260, whereas selecting the smallest initial rKL incurs $1.67$ and $0.89$ points of regret for the 0.6B and 1.7B students. 
At step 200, the reinforceable-fraction rule incurs $0.71$ points of regret for the 0.6B student and zero for the other two students, showing that its agreement with the best observed teacher also depends on the training horizon.

\begin{table}[htbp]
\centering
\small
\caption{Teacher-selection regret in accuracy points relative to
the best observed teacher at each horizon. Lower is better.
4B comparisons use mean outcomes and diagnostics across three runs.}
\label{tab:dynamics_selection}
\setlength{\tabcolsep}{6pt}
\renewcommand{\arraystretch}{1.1}
\begin{tabular}{lrrrrrr}
\toprule
& \multicolumn{3}{c}{Step 200}
& \multicolumn{3}{c}{Step 260} \\
\cmidrule(lr){2-4}\cmidrule(l){5-7}
Selector & 0.6B & 1.7B & 4B & 0.6B & 1.7B & 4B \\
\midrule
Largest teacher
& 1.35 & 2.67 & 3.44 & 1.67 & 0.89 & 2.92 \\
Smallest teacher
& 0.71 & 0.00 & 0.00 & 0.00 & 0.00 & 0.00 \\
Largest $\rho_0^+$
& 0.71 & 0.00 & 0.00 & 0.00 & 0.00 & 0.00 \\
Smallest initial rKL
& 1.35 & 2.67 & 0.00 & 1.67 & 0.89 & 0.00 \\
\bottomrule
\end{tabular}
\end{table}

In this grid, selecting the largest reinforceable fraction gives the same teacher choices as selecting the smallest teacher, so these results do not establish a selection advantage over teacher size alone. 
They do show that $\rho_0^+$ distinguishes useful pairings that minimum rKL fails to identify.
We therefore interpret $\rho_0^+$ as a diagnostic of the supervision encountered by a particular student, capturing the fraction of sampled response tokens with positive distillation advantages. 
This sign information complements the distributional discrepancy summarized by rKL, although neither statistic alone fully describes downstream task gains. 
These findings motivate our composition experiments, which examine how student preparation changes the supervision it receives from a fixed teacher.

\subsection{Teacher-Dependent Training Dynamics}
\label{app:dynamics:controls}

\begin{figure*}[!htbp]
\centering
\includegraphics[width=\textwidth]{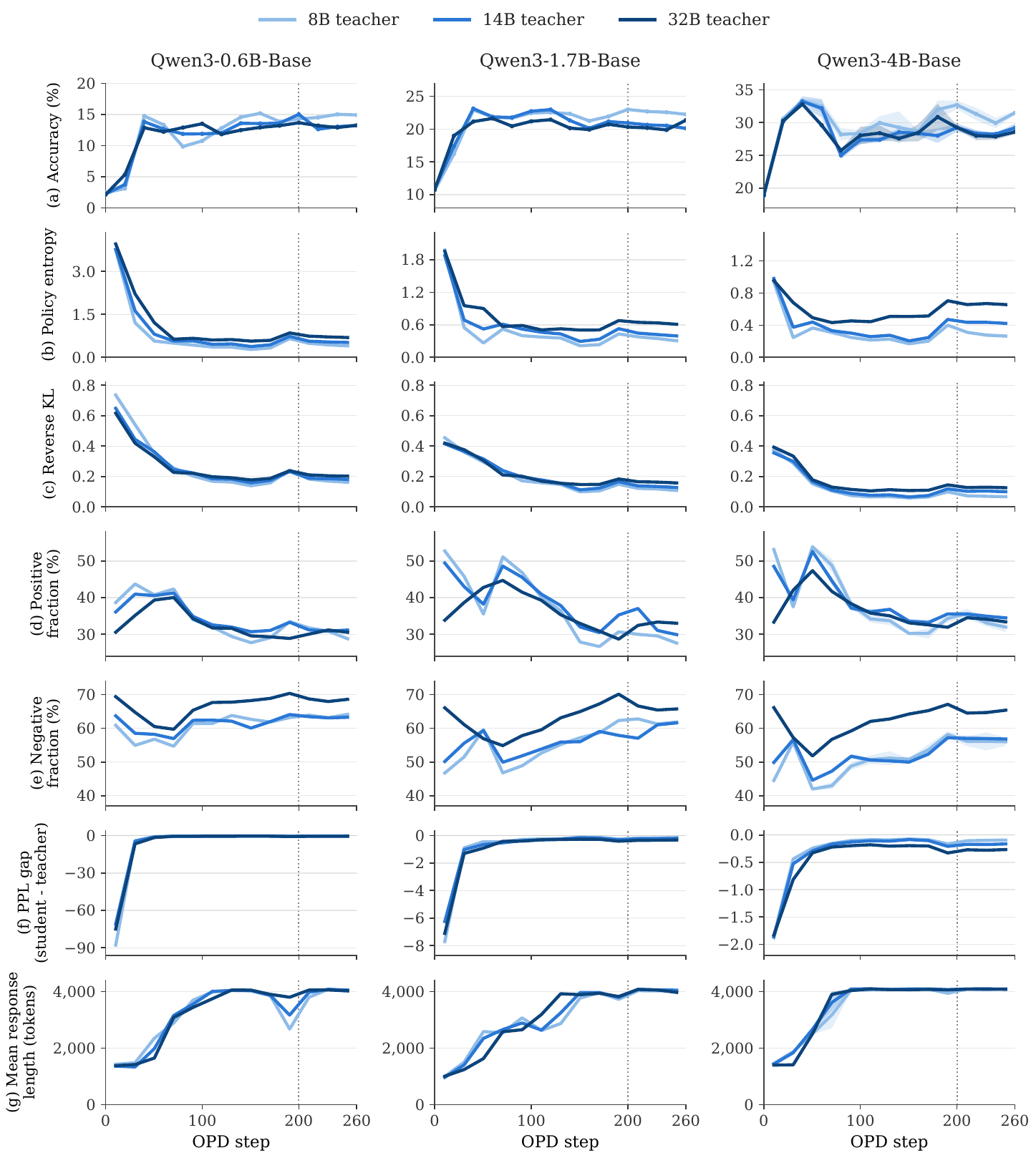}
\caption{OPD training dynamics across student and teacher scales.}
\label{fig:app_b_distillation_diagnostics}
\end{figure*}

We examine how teacher size shapes the evolution of distillation diagnostics for Qwen3-4B-Base.
\Cref{fig:app_b_distillation_diagnostics} summarizes three runs per teacher.
We normalize each run's reverse KL by its first logged value and average diagnostics over consecutive 20-step intervals before computing means and sample standard deviations across runs.

All three teachers show a pronounced early decrease in normalized reverse KL, accompanying the initial accuracy gains in \Cref{fig:app_b_learning_curves}c.
After this initial transition, normalized reverse KL remains lowest for the 8B teacher, intermediate for 14B, and highest for 32B.
The smaller teacher therefore permits a greater proportional reduction in the initial student--teacher discrepancy.
However, continued teacher matching does not consistently improve task performance: reverse KL remains low even when accuracy falls below its earlier maximum.
The reinforceable fraction follows a different pattern.
The 8B teacher initially provides the largest fraction of positive supervision and the 32B teacher the smallest, but these differences narrow and their ordering changes during training.
Thus, the persistent ordering in normalized reverse KL does not imply a persistent ordering in positive signal.
Together, these trajectories show that teacher matching, reinforceable supervision, and task performance capture distinct aspects of distillation.
Initialization diagnostics help characterize the starting pairing, while subsequent training should be assessed alongside task accuracy.

\begin{figure*}[t]
\centering
\includegraphics[width=\textwidth]{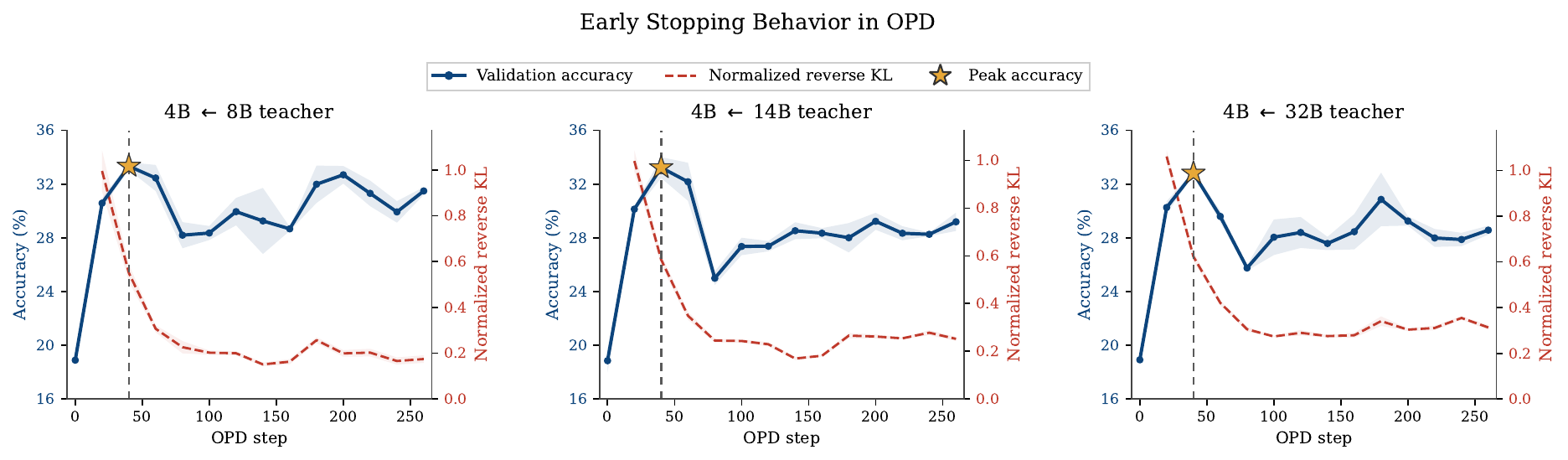}
\vspace{-6mm}
\caption{\textbf{Task accuracy and teacher matching follow different trajectories.}
Colored curves show mean benchmark accuracy for Qwen3-4B-Base, with sample-deviation bands across three runs per teacher.
Dashed gray curves show mean rKL normalized by each run's first logged value, using the right axis.
Dotted vertical lines mark the peak of each mean accuracy curve through step 200.}
\label{fig:stopping}
\end{figure*}

\subsection{OPD can lose task performance despite continued teacher matching.}
We observed that 4B student runs improve rapidly, but further teacher matching does not consistently preserve their early gains. Across the nine runs, stopping at step 40 gives $33.13\%$ mean accuracy, only $0.10$ points below the mean of their individual-run peaks through step 200. 
Continuing to step 200 instead yields $30.39\%$, a mean gap of $2.84$ points. 
The step-40 states therefore retain more accuracy with $80\%$ fewer optimizer updates than step 200. 
An exploratory $\kappa\leq0.6$ stopping rule selects step $48.9$ on average and gives $32.14\%$ accuracy, with a $1.09$-point gap and $75.6\%$ fewer updates than step 200. 
The fixed step-40 comparison performs better in these runs, while the threshold rule illustrates how a training-time diagnostic can be evaluated against a simple budget-based alternative. 
These comparisons are retrospective. 
Appendix~\ref{app:dynamics:stopping} provides the threshold sensitivity analysis and step accounting, including the extended step-260 reference. 
These results motivate evaluating whether OPD still improves task performance when allocating training to a subsequent stage.

\subsection{Early Stopping and Training-Budget Accounting}
\label{app:dynamics:stopping}
We compare fixed-step and KL-based stopping rules across the nine Qwen3-4B-Base runs to quantify how much task performance can be retained with fewer OPD updates. 
The analysis uses recorded evaluations at 20-step intervals through step 260. For each run, we measure the accuracy gap in percentage points between its selected evaluation and its highest observed accuracy within this horizon, then average these gaps across runs. 
A smaller gap indicates better retention of the observed peak. 
These \textit{retrospective comparisons} use evaluated policy states and do not assume that every state was retained as a saved checkpoint.

For the KL-based rule, let $d_r(s)$ denote run $r$'s logged sampled-token reverse KL at step $s$. 
We normalize it by the first logged value:
\begin{equation}
\kappa_r(s)=\frac{d_r(s)}{d_r(1)},
\end{equation}
and stop at the first evaluated step $s\geq20$ satisfying $\kappa_r(s)\leq\tau$, using step 260 if no crossing occurs. The rule uses the unsmoothed statistic at each evaluation step. 
We compare $\tau\in\{0.4,0.5,0.6,0.7,0.8\}$, where a smaller threshold requires a larger reduction in rKL before stopping.

\begin{table}[htbp]
\centering
\small
\caption{Stopping comparisons averaged over nine Qwen3-4B-Base runs. Gap is measured from each run's observed peak through step 260. Update savings use step 260 as the reference.}
\label{tab:dynamics_stopping}
\setlength{\tabcolsep}{4pt}
\begin{tabular}{lrrrr}
\toprule
Stopping rule & Mean step & Accuracy (\%) & Gap (pp)
& Updates saved (\%) \\
\midrule
Fixed step 40  & 40.0  & 33.13 & 0.10 & 84.6 \\
Fixed step 60  & 60.0  & 31.41 & 1.82 & 76.9 \\
Fixed step 100 & 100.0 & 27.92 & 5.31 & 61.5 \\
Fixed step 200 & 200.0 & 30.39 & 2.84 & 23.1 \\
Fixed step 260 & 260.0 & 29.75 & 3.48 & 0.0 \\
\midrule
$\tau=0.4$ & 66.7 & 30.13 & 3.11 & 74.4 \\
$\tau=0.5$ & 60.0 & 31.41 & 1.82 & 76.9 \\
$\tau=0.6$ & 48.9 & 32.14 & 1.09 & 81.2 \\
$\tau\in\{0.7,0.8\}$ & 40.0 & 33.13 & 0.10 & 84.6 \\
\bottomrule
\end{tabular}
\end{table}

\Cref{tab:dynamics_stopping} shows that a fixed stop at step 40 achieves $33.13\%$ mean accuracy, only $0.10$ points below the mean of the individual-run peaks ($33.23\%$). 
Continuing to steps 200 and 260 instead yields $30.39\%$ and $29.75\%$. 
The step-40 state therefore retains more accuracy with $80\%$ fewer updates than step 200 and $84.6\%$ fewer than step 260. The $\tau=0.6$ rule stops at step $48.9$ on average and achieves $32.14\%$ accuracy, with a mean gap of $1.09$ points. Its effect varies across teachers: the gaps are $0.04$, $0.00$, and $3.23$ points for the 8B, 14B, and 32B teachers, respectively.
Thresholds $0.7$ and $0.8$ select step 40 in every run and match the fixed-step result, while lower thresholds wait longer and produce larger mean gaps in this sweep. 
These comparisons identify an effective early stopping opportunity in the observed runs, but do not independently validate a stopping step or threshold for new student--teacher pairs.

With 64 prompts per update and four responses per prompt, step 40 requires 2,560 prompt presentations and 10,240 sampled responses. 
The $\tau=0.6$ rule averages approximately 3,129 prompt presentations and 12,516 responses, compared with 16,640 and 66,560 at step 260. 
These counts include repeated presentations of a question. 
Reported savings concern optimizer updates and sampled responses, while token and compute costs also depend on response length, teacher scoring, optimization, and evaluation. 
The practical implication is to assess whether further OPD still improves task accuracy when deciding how much training to allocate before a subsequent stage.

\section{Student Warm-Up Before Distillation}
\label{app:student_warmup}

This appendix provides additional implementation details and analyses for the student warm-up experiments in \S\ref{subsec:student_warmup}. 
We describe the SFT and RLVR warm-up procedures, the shared downstream OPD setup, and the effect of SFT duration on subsequent distillation.

\subsection{SFT Warm-Up}
\label{app:student_warmup_sft}

We warm up Qwen3-4B-Base with full-parameter SFT on correct trajectories generated by Qwen3-14B for OpenThoughts-Math prompts.
We sweep the resulting student checkpoints at $K\in\{25,75,150,250,348\}$ training steps. 
Each checkpoint independently initializes a downstream OPD run against the same fixed Qwen3-14B teacher, allowing us to isolate how the amount of SFT before distillation changes the subsequent optimization. 
Table~\ref{tab:student_warmup_sft} reports the complete training configuration.

\begin{table}[!htbp]
\centering
\small
\caption{Student SFT warm-up hyperparameters.}
\label{tab:student_warmup_sft}
\renewcommand{\arraystretch}{1.12}
\begin{tabular}{@{}ll@{}}
\toprule
\textbf{Setting} & \textbf{Student SFT Warm-Up} \\
\midrule
Starting student & Qwen3-4B-Base \\
Solution generator & Qwen3-14B \\
Prompt source & OpenThoughts-Math \\
Supervision & Fixed teacher-generated solutions \\
Update scope & Full-parameter SFT \\
Training backend / GPUs & FSDP / 8 \\
Microbatch per GPU & 4 \\
Learning rate & $10^{-5}$ \\
Training duration & 3 epochs \\
Maximum sequence length & 14,336 tokens \\
Truncation & Right \\
Sequence parallel size & 2 \\
Padding removal & Enabled \\
Save / validation interval & 25 / 25 steps \\
Validation criterion & SFT loss \\
\bottomrule
\end{tabular}
\end{table}

\subsection{RLVR Warm-Up}
\label{app:student_warmup_rlvr}

For the RLVR warm-up control, we instead train Qwen3-4B-Base on OpenThoughts-Math with GRPO and a binary verifiable outcome reward. 
For each prompt, the student samples eight responses and receives reward according to solution correctness. 
Unlike SFT warm-up, this stage improves the student directly from task reward and does not use the downstream distillation teacher as supervision.

Selected RLVR checkpoints are subsequently distilled from the same fixed Qwen3-14B teacher on DAPO-Math-17K. 
The OPD step counter is restarted at this handoff, so warm-up optimization and downstream distillation are treated as separate stages. Table~\ref{tab:student_warmup_rlvr} reports the RLVR configuration.

\begin{table}[!htbp]
\centering
\small
\caption{Student RLVR warm-up hyperparameters.}
\label{tab:student_warmup_rlvr}
\renewcommand{\arraystretch}{1.10}
\begin{tabular}{@{}ll@{}}
\toprule
\textbf{Setting} & \textbf{Student RLVR Warm-Up} \\
\midrule
Starting student & Qwen3-4B-Base (Thinking Disabled) \\
Algorithm & GRPO \\
Warm-up data & OpenThoughts-Math \\
Reward & Verifiable outcome (binary) \\
Update scope & Full-parameter training \\
Prompt / PPO minibatch size & 64 / 64 \\
Responses per prompt & 8 \\
Rollout temperature & 1.0 \\
Prompt / response limit & 2,048 / 8,192 tokens \\
Learning rate / warm-up & $10^{-6}$ / 10 steps \\
Weight decay / gradient clip & 0.01 / 1.0 \\
Lower / upper clipping ratio & 0.20 / 0.28 \\
Rollout importance correction & Token level; threshold 2.0 \\
KL reward / KL loss & Disabled / disabled \\
Entropy coefficient / loss aggregation & 0 / token mean \\
Data shuffling / overlong-prompt filtering & Disabled / enabled \\
Training schedule & 1 epoch; no explicit step cap \\
Save / evaluation interval & 20 / 20 steps \\
Evaluation temperature / top-$p$ & 0.6 / 0.95 \\
Evaluation responses / top-$k$ & 4 / disabled \\
Training backend / GPUs & FSDP / 8 \\
Rollout backend / tensor parallelism & vLLM / 1 \\
Sequence parallel size & 1 \\
Dynamic token budget per GPU & 32,768 \\
Padding removal / gradient checkpointing & Enabled / enabled \\
\bottomrule
\end{tabular}
\end{table}

\subsection{Downstream OPD Setup}
\label{app:student_warmup_opd}

All warm-up checkpoints are evaluated under the same downstream OPD recipe. 
We distill from a fixed Qwen3-14B teacher on DAPO-Math-17K using the sampled-token reverse-KL policy-gradient objective described in \S\ref{sec:background}. 
Student initialization is therefore the only quantity changed across the SFT checkpoint sweep. We evaluate every $20$ OPD steps using four responses per problem at temperature $0.6$ and top-$p=0.95$. 
Reported accuracy is mean@4 across AIME24, AIME25, AMC23, GPQA, MATH-500, Minerva, and Olympiad-Bench. Table~\ref{tab:student_warmup_opd} gives the complete downstream configuration.

\begin{table}[!htbp]
\centering
\small
\caption{Downstream OPD hyperparameters for the SFT warm-up sweep.}
\label{tab:student_warmup_opd}
\renewcommand{\arraystretch}{1.10}
\begin{tabular}{@{}ll@{}}
\toprule
\textbf{Setting} & \textbf{Downstream OPD} \\
\midrule
Student initialization & Selected SFT checkpoint \\
Teacher & Qwen3-14B \\
Training data & DAPO-Math-17K \\
Objective & Sampled-token reverse KL policy gradient \\
Task rewards / auxiliary KL loss & Disabled / disabled \\
Optimizer & AdamW \\
Learning rate / warm-up & $10^{-6}$ / 10 steps \\
LR schedule / Adam betas & Constant / $(0.9, 0.999)$ \\
Weight decay / gradient clip & 0.01 / 1.0 \\
Prompt / PPO minibatch size & 64 / 64 \\
PPO epochs / clipping ratio & 1 / 0.2 \\
Responses per prompt & 4 \\
Rollout temperature / top-$p$ & 1.0 / 1.0 \\
Prompt / response limit & 2,048 / 8,192 tokens \\
Loss aggregation / entropy coefficient & Token mean / 0 \\
Distillation coefficient / configured top-$k$ & 1 / 64 \\
Loss / log-probability clamps & 10 / $-10$ \\
Training duration & 1 epoch; 200-step cap \\
Save / evaluation interval & 50 / 20 steps \\
Evaluation temperature / top-$p$ & 0.6 / 0.95 \\
Evaluation responses / top-$k$ & 4 / disabled \\
Data shuffling / overlong-prompt filtering & Disabled / enabled \\
Student / teacher inference dtype & bfloat16 / bfloat16 \\
Student / teacher GPUs & 4 / 4 \\
Student rollout / teacher TP & 1 / 2 \\
Student / teacher GPU memory fraction & 0.70 / 0.85 \\
Dynamic token budget per GPU & 24,576 \\
GPU type & NVIDIA H200 \\
\bottomrule
\end{tabular}
\end{table}

\newpage
\subsection{Sensitivity to SFT Warm-Up Duration}
\label{app:student_warmup_results}

\begin{wrapfigure}{r}{0.39\textwidth}
\vspace{-10pt}
\centering
\includegraphics[width=\linewidth]{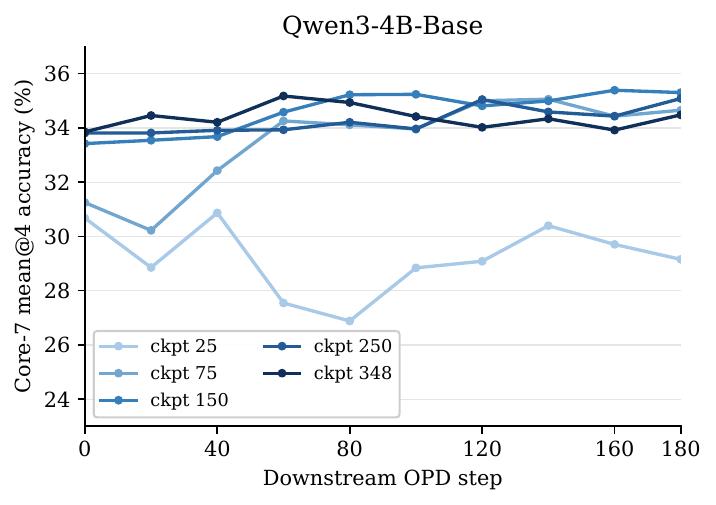}
\caption{\textbf{Downstream OPD trajectories from different SFT warm-up checkpoints.}}
\label{fig:student_warmup_sweeps}
\vspace{-8pt}
\end{wrapfigure}
Figure~\ref{fig:student_warmup_sweeps} shows that the benefit of SFT warm-up appears quickly and then saturates. The downstream OPD peak increases from $30.87\%$ at $K=25$ to $35.06\%$ at $K=75$ and $35.39\%$ at $K=150$. Extending SFT further gives peaks of $35.08\%$ at $K=250$ and $35.18\%$ at $K=348$. Thus, most of the downstream gain is already realized by $K=75$--$150$, and additional SFT does not raise the OPD ceiling. The longer warm-ups also largely preserve their peak performance throughout OPD. In particular, the $K=150$ run finishes at $35.30\%$, only $0.09$ points below its best observed accuracy, while the $K=250$ run finishes at its peak. By contrast, the minimally warmed $K=25$ student both reaches a lower maximum and exhibits a larger subsequent drop. These trajectories support using a short SFT phase to prepare the student rather than continuing SFT until convergence.

\begin{table}[!htbp]
\centering
\small
\caption{Downstream OPD results across SFT warm-up checkpoints.}
\label{tab:student_warmup_results}
\renewcommand{\arraystretch}{1.12}
\begin{tabular}{@{}rrrrrr@{}}
\toprule
\textbf{SFT Checkpoint $K$} &
\textbf{Initial (\%)} &
\textbf{Peak (\%)} &
\textbf{Peak $t$} &
\textbf{Final (\%)} &
\textbf{Drop} \\
\midrule
25  & 30.68 & 30.87 & 40  & 29.15 & 1.71 \\
75 & 31.25 & 35.06 & 140 & 34.65 & 0.41 \\
150 & 33.42 & 35.39 & 160 & 35.30 & 0.09 \\
250 & 33.81 & 35.08 & 180 & 35.08 & 0.00 \\
348 & 33.85 & 35.18 & 60  & 34.48 & 0.70 \\
\bottomrule
\end{tabular}
\end{table}

\subsection{Initialization Diagnostics}
\label{app:student_warmup_diagnostics}

The checkpoint sweep also shows that continued SFT keeps changing the student even after its downstream benefit has saturated. Student entropy at the beginning of OPD decreases from $0.2940$ at $K=25$ to $0.1950$ at $K=150$ and $0.1507$ at $K=250$, yet the strongest downstream OPD result is already obtained at $K=150$. Lower initialization entropy therefore does not monotonically translate into better distillation.

A similar pattern appears in the initial teacher--student reverse KL. It falls from $0.1417$ at $K=25$ to its minimum of $0.1028$ at $K=150$, but further SFT does not produce additional downstream gains. Together with the validation-loss curve in Figure~\ref{fig:warmup}d, this places the useful warm-up region around $K=75$--$150$: the first validation-loss plateau coincides with saturation in downstream OPD. We use this as a practical signal of diminishing returns rather than as a universal stopping rule.

\begin{table}[!htbp]
\centering
\small
\caption{Student diagnostics at the beginning of downstream OPD.}
\label{tab:student_warmup_diagnostics}
\renewcommand{\arraystretch}{1.12}
\begin{tabular}{@{}rrrr@{}}
\toprule
\textbf{SFT Checkpoint $K$} &
\textbf{OPD Step $t$} &
\textbf{Student Entropy} &
\textbf{Reverse KL} \\
\midrule
25  & 1 & 0.2940 & 0.1417 \\
75 & 1 & 0.2559 & 0.1227 \\
150 & 1 & 0.1950 & 0.1028 \\
250 & 1 & 0.1507 & 0.1153 \\
348 & 1 & 0.1580 & 0.1117 \\
\bottomrule
\end{tabular}
\end{table}

\section{Teacher Adaptation Before Distillation}
\label{app:teacher_adaptation}

This appendix provides the details of teacher preparation and the corresponding downstream OPD settings for \S\ref{subsec:teacher_adaptation}, along with additional results and discussion for interested readers.

\subsection{Experimental Setup}
\label{app:teacher_adaptation_setup}

We compare Qwen3-14B teachers adapted using either SFT on positive self-generated rollouts (\citep{zelikman2022star}) or RLVR on OpenThoughts3-Math. Each teacher then distills a fresh Qwen3-4B-Base student on DAPO-Math-17K under identical downstream settings, including student initialization, OPD objective, learning rate, batch size, sampling, and evaluation. Table~\ref{tab:teacher_sft_config} reports the SFT teacher-adaptation hyperparameters, while Table~\ref{tab:teacher_opd_config} reports the RLVR hyperparameters, separating teacher preparation from downstream student training.

\subsection{Teacher Adaptation with SFT}
\label{app:teacher_adaptation_sft}

For the SFT-adapted teacher, we start from Qwen3-14B with thinking disabled and construct supervision directly from its own rollouts on OpenThoughts3-Math. For each prompt, we sample up to four independent responses and retain the prompt only if at least one rollout is verified as correct by the task reward. Incorrect generations are discarded, and sampling is repeated, up to the four-generation budget, until a verified-correct solution is obtained. This produces approximately $16$K unique prompts paired with verified self-generated solutions, matching the number of unique training prompts used for RLVR teacher adaptation and thereby controlling for adaptation-data scale. We then perform full-parameter SFT on this fixed dataset for three epochs using AdamW with a learning rate of $10^{-6}$ and a maximum sequence length of $14{,}336$ tokens. Selected checkpoints from this training trajectory are subsequently used as teachers for downstream OPD. 

\begin{table}[!htbp]
\centering
\small
\caption{SFT teacher training hyperparameters.}
\label{tab:teacher_sft_config}
\renewcommand{\arraystretch}{1.12}
\begin{tabular}{@{}ll@{}}
\toprule
\textbf{Setting} & \textbf{SFT Teacher} \\
\midrule
Starting teacher & Qwen3-14B (Thinking Disabled) \\
Adaptation data & OpenThoughts3-Math \\
SFT format & Full-parameter \\
Supervision & Fixed, verified \textbf{self}-generated solutions \\
Unique prompts in batch & 64 \\
Learning rate & $10^{-6}$ \\
Maximum sequence length & 14,336 tokens \\
Truncation & Right \\
Training schedule & 3 epochs \\
Optimizer & AdamW \\
Training backend & FSDP \\
GPUs & 8 \\
\bottomrule
\end{tabular}
\end{table}

\subsection{RLVR Teacher Adaptation}
\label{app:teacher_adaptation_rlvr}

For the RLVR-adapted teacher, we start from the same Qwen3-14B model and use the same approximately $16$K OpenThoughts3-Math training prompts considered for SFT adaptation in \S\ref{app:teacher_adaptation_sft}. This controls the underlying adaptation data and isolates the effect of the teacher-training objective. We train the teacher with GRPO using the common RLVR optimization and rollout recipe adopted throughout this work: for each prompt, we sample eight responses at temperature $1.0$ and optimize against a binary verifiable-outcome reward indicating whether the generated solution is correct. Training uses batches of $64$ unique prompts, a learning rate of $10^{-6}$ with a $10$-step warm-up, and a single pass over the adaptation set. We disable both KL regularization and entropy bonuses and otherwise retain the RLVR optimization, clipping, importance-correction, and rollout settings used in our main experiments. To study how progressive RLVR adaptation of the teacher affects subsequent distillation, we save checkpoints every $20$ training steps and select five checkpoints along the training trajectory for downstream OPD evaluation. Each checkpoint is used to distill a freshly initialized Qwen3-4B-Base student under the same downstream OPD configuration, allowing differences in student performance to be attributed to the teacher's degree of RLVR adaptation. Table~\ref{tab:teacher_rlvr_config} reports the complete RLVR teacher-training configuration.

\begin{table}[!htbp]
\centering
\small
\caption{RLVR teacher training hyperparameters.}
\label{tab:teacher_rlvr_config}
\renewcommand{\arraystretch}{1.1}
\begin{tabular}{@{}ll@{}}
\toprule
\textbf{Setting} & \textbf{RLVR-Adapted Teacher} \\
\midrule
Starting teacher & Qwen3-14B \\
Algorithm & GRPO \\
Adaptation data & OpenThoughts-Math \\
Reward & Verifiable outcome (binary) \\
Unique prompts in batch & 64 \\
Responses per prompt / temperature & 8 / 1.0 \\
Prompt / response limit & 2,048 / 8,192 tokens \\
Learning rate / warm-up & $10^{-6}$ / 10 steps \\
Weight decay / gradient clip & 0.01 / 1.0 \\
Lower / upper clipping ratio & 0.20 / 0.28 \\
Rollout importance correction & Token level w/. threshold 2.0 \\
KL loss & Disabled \\
Entropy coefficient / loss aggregation & 0 / token mean \\
Data shuffling / overlong-prompt filtering & Disabled / enabled \\
Training schedule & 1 epoch; no explicit step cap \\
Save / evaluation interval & 20 / 20 steps \\
Evaluation temperature / top-$p$ & 0.6 / 0.95 \\
Evaluation samples / top-$k$ & 4 / disabled \\
Training GPUs / rollout TP & 8 / 1 \\
Rollout backend / GPU memory fraction & vLLM / 0.60 \\
Actor parameter / optimizer offload & Enabled / enabled \\
\bottomrule
\end{tabular}
\end{table}

\subsection{Downstream OPD Configuration and Evaluation}
\label{app:teacher_adaptation_opd}

Table~\ref{tab:teacher_opd_config} shows our downstream OPD hyperpaarameters after both SFT and RLVR teacher. Its uses sampled-token reverse-KL policy gradients on student-generated
responses, with task rewards and auxiliary KL regularization disabled.
The student uses the non-thinking chat-template setting throughout OPD.
The SFT and RLVR teacher runs both runs for $200$ steps and validated in every 20 steps. During evaluaton, for each checkpoint, we sample four responses per evaluation question at
temperature $0.6$ and top-$p=0.95$. We compute accuracy as the unweighted
mean of the seven benchmark-specific mean@4 scores: AIME24, AIME25, AMC23,
MATH-500, Minerva, Olympiad-Bench, and GPQA. Thus, each benchmark receives
equal weight, and mean@4 averages correctness over the four responses.

\begin{table}[!htbp]
\centering
\small
\caption{Downstream OPD training hyperparameters shared across SFT- and RLVR-adapted teachers.}
\label{tab:teacher_opd_config}
\begin{tabular}{@{}ll@{}}
\toprule
\textbf{Setting} & \textbf{SFT/RLVR Teacher} \\
\midrule
Student & Qwen3-4B-Base \\
Training data & DAPO-Math-17K \\
Objective & Sampled-token reverse KL policy gradient \\
Task rewards / auxiliary KL loss & Disabled / disabled \\
Optimizer & AdamW \\
Learning rate / warm-up & $10^{-6}$ / 10 steps \\
LR schedule / Adam betas & Constant / $(0.9, 0.999)$ \\
Weight decay / gradient clip & 0.01 / 1.0 \\
Prompt size & 64 \\
PPO epochs / clipping ratio & 1 / 0.2 \\
Rollout temperature / top-$p$ & 1.0 / 1.0 \\
Prompt / response limit & 2,048 / 8,192 tokens \\
Loss aggregation / entropy coefficient & Token mean / 0 \\
Distillation coefficient / configured top-$k$ & 1 / 64 \\
Loss / log-probability clamps & 10 / $-10$ \\
Configured epochs / step cap & 1 / 200 \\
Save / evaluation interval & 20 / 20 steps \\
Eval temperature / top-$p$ & 0.6 / 0.95 \\
Evaluation responses / top-$k$ & 4 / disabled \\
Student FSDP / inference dtype & bfloat16 / bfloat16 \\
Student rollout / teacher TP & 1 / 2 \\
Student / teacher GPU memory fraction & 0.70 / 0.85 \\
Dynamic token budget per GPU & 24,576 \\
GPU type & 8$\times$ NVIDIA B200 \\
\bottomrule
\end{tabular}
\end{table}

\subsection{Teacher Selection at Fixed Model Size}
\label{app:teacher_adaptation_diagnostics}

\begin{wraptable}{r}{0.52\textwidth}
\centering
\small
\caption{Initial diagnostics and step-200 accuracy for Qwen3-1.7B-Base with RLVR-adapted Qwen3-14B teachers. Each configuration uses one run.}
\label{tab:teacher_adaptation_diagnostics}
\setlength{\tabcolsep}{3pt}
\renewcommand{\arraystretch}{1.1}
\begin{tabular*}{\linewidth}{@{\extracolsep{\fill}}lrrr@{}}
\toprule
Teacher $K$ & rKL$_0$ & $\rho_0^+$ (\%) & $A_{200}$ (\%) \\
\midrule
50  & 0.505 & 47.23 & 27.41 \\
100 & \textbf{0.476} & 52.40 & 28.00 \\
150 & 0.555 & 49.08 & 29.16 \\
200 & 0.530 & \textbf{52.97} & \textbf{29.77} \\
\bottomrule
\end{tabular*}
\end{wraptable}
To examine whether initialization diagnostics distinguish useful teachers beyond parameter count, we compare four RLVR-adapted Qwen3-14B checkpoints using Qwen3-1.7B-Base. 
Each checkpoint distills a freshly initialized student under identical downstream OPD settings. \Cref{tab:teacher_adaptation_diagnostics} reports reverse KL and reinforceable fraction at the first logged training step, together with accuracy at OPD step $200$. All runs use rollout seed $42$, and checkpoint labels follow \Cref{fig:teacher_adaptation_sweeps}. 

Selecting maximum $\rho_0^+$ identifies $K=200$, which achieves the highest accuracy in this comparison at $29.77\%$. Minimum reverse KL instead selects $K=100$, yielding $28.00\%$ and incurring $1.77$ percentage points of selection regret. Thus, the teacher closest to the student under initial reverse KL does not provide the most effective supervision at the evaluated training budget. Because all candidates share the same parameter count, this comparison illustrates the diagnostic value of $\rho_0^+$ beyond teacher size. Its sign information helps distinguish useful checkpoints that minimum divergence does not identify, supporting its use as a complementary measure of the initial student--teacher pairing in this setting.

\subsection{Additional Results}
\label{app:teacher_adaptation_results}

\begin{figure*}[!htbp]
\centering
\includegraphics[width=\textwidth]{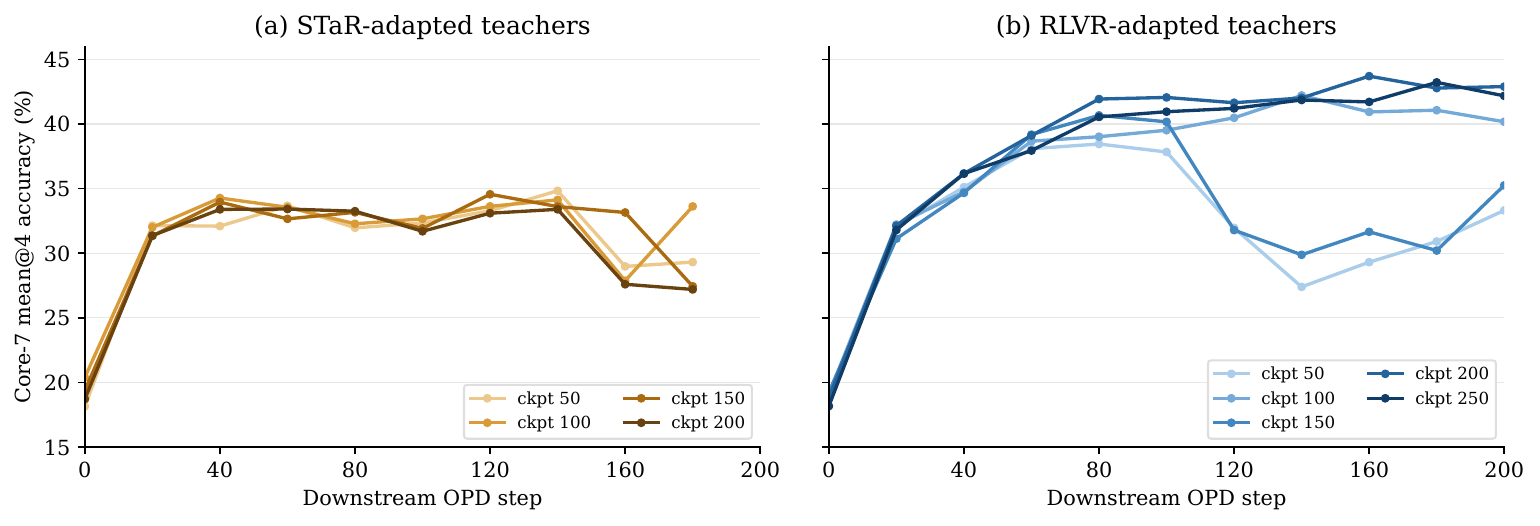}
\caption{Downstream OPD performances across SFT and RLVR adapted teacher checkpoint sweeps.}
\label{fig:teacher_adaptation_sweeps}
\end{figure*}

Figure~\ref{fig:teacher_adaptation_sweeps} shows the full downstream OPD trajectories for individual teacher checkpoints. With SFT adaptation, the runs remain closely aligned through approximately OPD step $140$: more heavily adapted teachers converge to nearly the same accuracy as earlier checkpoints, with only small differences, before several runs decline afterward. This is consistent with SFT changing the teacher policy without materially increasing the amount of task capability available for distillation.

RLVR exhibits the opposite trend. Up to roughly OPD step $100$, stronger RLVR teacher checkpoints generally produce progressively higher student accuracy. Beyond this point, however, the less-adapted checkpoints, particularly $K=50$ and $K=100$, become substantially less stable, while the more strongly adapted teachers retain higher performance. Among the later checkpoints, $K=200$ reaches the strongest performance, slightly ahead of $K=250$, with both outperforming $K=150$, which in turn improves over the earlier checkpoints. These trajectories suggest a capability--stability trade-off in teacher adaptation: increasing RLVR initially strengthens the transferable learning signal, but the benefit saturates once teacher capability plateaus. In practice, this suggests stopping teacher RLVR near its capability plateau and selecting the teacher checkpoint using downstream OPD validation rather than simply using the final checkpoint.

\section{Downstream RLVR Initialization}
\label{app:stage_ordering}

In this section of Appendix, we detail the stage-ordering comparison in \S\ref{subsec:stage_ordering}: SFT and OPD prepare a student for the same downstream RLVR procedure. 
We report the initialization and RLVR configurations separately, followed by additional results across student scales and model families.

\subsection{Experimental Setting}
\label{app:stage_ordering_protocol}

For each student, the two initialization methods use the same starting model and the same fixed teacher. 
Qwen3 experiments pair Qwen3-Base students (1.7B and 4B) with Qwen3-14B and the Qwen2.5 experiments pair Qwen2.5-1.5B-Base with Qwen2.5-14B-Instruct. 
SFT learns from fixed, verified teacher-generated solutions, whereas OPD obtains teacher feedback on student-sampled responses. 
Both preparatory stages use OpenThoughts3-Math data. 
Their checkpoints then initialize GRPO on DAPO-Math-17K, with the downstream settings held fixed across the two arms within each student comparison.
The initialization launchers specify a batch size of $64$ and learning rate $10^{-6}$. 
SFT uses one stored solution per training example, while OPD samples four responses per prompt following but the number of unique training samples each see is same following \cite{li2026rethinking} and \cite{zhu2026many}.

\subsection{SFT Initialization}
\label{app:stage_ordering_sft}

Table~\ref{tab:stage_ordering_sft} gives the SFT configuration, which is shared by every student within a model family.
The SFT data were built from the teacher that each family's students are later distilled from: Qwen3-14B for the Qwen3 students and Qwen2.5-14B-Instruct for the Qwen2.5 students, so that the comparison covers both a range of student sizes and a second architecture.
We took the prompts from OpenThoughts3-Math, sampled one solution per prompt from the corresponding teacher, graded it against the gold answer, and kept only the prompts whose solution was correct.
The construction script retains one verified teacher solution per solvable question and holds out 256 of these examples for SFT validation.
It passes the full retained question pool, including those 256, to OPD as its prompt set.
The two stages therefore see the same unique training prompts in the same order, and differ only in that SFT withholds the validation subset.

\begin{table}[!htbp]
\centering
\caption{
\textbf{SFT initialization settings.} 
The reported comparison on Qwen3-1.7B-Base, Qwen3-4B-Base, and Qwen2.5-1.5B students.
}
\label{tab:stage_ordering_sft}
\small
\setlength{\tabcolsep}{4pt}
\renewcommand{\arraystretch}{1.12}
\begin{tabularx}{\linewidth}{@{}>{\raggedright\arraybackslash}p{0.32\linewidth}XX@{}}
\toprule
Setting & Qwen3 family & Qwen2.5 family \\
\midrule
Supported base students & 0.6B, 1.7B, 4B & 1.5B \\
Teacher & Qwen3-14B & Qwen2.5-14B-Instruct \\
Training data & \multicolumn{2}{l}{OpenThoughts3-Math} \\
Supervision & \multicolumn{2}{l}{Fixed, verified teacher-generated solutions} \\
Unique prompts in batch & 64 & 64 \\
Microbatch per GPU & 4 & 4 \\
Learning rate & $10^{-6}$ & $10^{-6}$ \\
Maximum sequence length & 6,145 tokens & 6,145 tokens \\
Truncation & Right & Right \\
Training schedule & 1 epoch; 250-step cap & 1 epoch; 250-step cap \\
Save / validation interval & 25 / 25 steps & 25 / 25 steps \\
Training backend & FSDP & FSDP \\
Student GPUs & 8 & 8 \\
Sequence parallel size & 2 & 2 \\
\bottomrule
\end{tabularx}
\end{table}

\subsection{OPD Initialization}
\label{app:stage_ordering_opd}
OPD needs no teacher rollouts. The student generates the trajectories, and the teacher, which is larger and more capable, scores every token of them, so the stage requires only a verifiable prompt set with gold answers.
We use the same OpenThoughts3-Math questions as the SFT stage, in the same number and the same order, so that at every optimizer step the two stages have consumed the same unique prompts.
Table~\ref{tab:stage_ordering_opd} gives the OPD configuration.
The student samples responses at temperature $1.0$ and learns from the fixed teacher through the sampled-token reverse-KL policy-gradient objective of Equation~\ref{eq:adv}.
The OPD checkpoints shown in the figures use the same step sets $K$ listed above for each student, so that the comparison covers several hand-off points.

\begin{table}[!htbp]
\centering
\caption{
\textbf{OPD initialization settings. }
All students use the same optimization and sampling settings, the teacher and initialization data follow the model family.}
\label{tab:stage_ordering_opd}
\small
\setlength{\tabcolsep}{4pt}
\renewcommand{\arraystretch}{1.12}
\begin{tabularx}{\linewidth}{@{}>{\raggedright\arraybackslash}p{0.32\linewidth}XX@{}}
\toprule
Setting & Qwen3 family & Qwen2.5 family \\
\midrule
Supported base students & 0.6B, 1.7B, 4B & 1.5B, 3B \\
Teacher & Qwen3-14B & Qwen2.5-14B-Instruct \\
Training data & \multicolumn{2}{l}{OpenThoughts3-Math} \\
Distillation objective & \multicolumn{2}{l}{Sampled-token policy gradient reverse-KL } \\
Task rewards & Disabled & Disabled \\
Prompt / PPO minibatch size & 64 / 64 & 64 / 64 \\
Responses per prompt & 4 & 4 \\
Rollout temperature & 1.0 & 1.0 \\
Prompt / response limit & 2,048 / 8,192 tokens & 2,048 / 8,192 tokens \\
Learning rate & $10^{-6}$ & $10^{-6}$ \\
LR warm-up & 10 steps & 10 steps \\
Weight decay / gradient clip & 0.01 / 1.0 & 0.01 / 1.0 \\
Clipping ratio & 0.2 & 0.2 \\
Loss aggregation & Token mean & Token mean \\
Auxiliary KL / entropy loss & Disabled / 0 & Disabled / 0 \\
Configured distillation top-$k$ & 64 & 64 \\
Loss / log-probability clamps & 10.0 / $-10.0$ & 10.0 / $-10.0$ \\
Training schedule & 1 epoch; 250-step cap & 1 epoch; 250-step cap \\
Save / evaluation interval & 25 / 25 steps & 25 / 25 steps \\
Student / teacher GPUs & 4 / 4 & 4 / 4 \\
Rollout / teacher tensor parallelism & 1 / 1 & 1 / 1 \\
Dynamic token budget per GPU & 24,576 & 24,576 \\
\bottomrule
\end{tabularx}
\end{table}

\subsection{Common Downstream RLVR Stage}
\label{app:stage_ordering_rlvr}
Each selected checkpoint is exported to Hugging Face format and loaded as the initial policy for a separate GRPO run. 
Table~\ref{tab:stage_ordering_rlvr} applies to both initialization arms. 
RLVR stage runs for one epoch over the processed DAPO-Math-17K data with no explicit step cap, so that every run covers the full dataset and each training prompt is seen by the optimizer exactly once.
Evaluation samples four responses per question at temperature $0.6$ and top-$p=0.95$, with top-$k$ filtering disabled on Qwen3 series and use temperature $0.7$ for Qwen2.5 followig their suggested HuggingFace decoding parameters. 
Similar to \S\ref{sec:dynamics}, we report held-out \texttt{core-7} mean@4 accuracy on seven core reasoning benchmarks covering math and general reasoning: AIME24, AIME25, AMC23, MATH-500, Minerva, Olympiad-Bench, and GPQA.

\begin{table}[!htbp]
\centering
\caption{
Common Downstream RLVR settings for SFT and OPD-initialized students.
The reported horizons reflect the displayed curves, while the launchers specify an epoch-based training schedule.
}
\label{tab:stage_ordering_rlvr}
\small
\setlength{\tabcolsep}{4pt}
\renewcommand{\arraystretch}{1.12}
\begin{tabularx}{\linewidth}{@{}>{\raggedright\arraybackslash}p{0.32\linewidth}XX@{}}
\toprule
Setting & Qwen3 family & Qwen2.5 family \\
\midrule
Algorithm & GRPO & GRPO \\
Training data & DAPO-Math-17K & DAPO-Math-17K \\
Reward & \multicolumn{2}{l}{Verifiable outcome reward (\texttt{ttrl\_math.reward\_func})} \\
Prompt / PPO minibatch size & 64 / 64 & 64 / 64 \\
Responses per prompt & 8 & 8 \\
Rollout temperature & 1.0 & 1.0 \\
Prompt / response limit & 2,048 / 8,192 tokens & 2,048 / 8,192 tokens \\
Learning rate / warm-up & $10^{-6}$ / 10 steps & $10^{-6}$ / 10 steps \\
Weight decay / gradient clip & 0.01 / 1.0 & 0.01 / 1.0 \\
Lower / upper clipping ratio & 0.20 / 0.28 & 0.20 / 0.28 \\
Rollout importance correction & Token level; threshold 2.0 & Token level; threshold 2.0 \\
KL reward / KL loss & Disabled / disabled & Disabled / disabled \\
Entropy coefficient & 0 & 0 \\
Loss aggregation & Token mean & Token mean \\
Data shuffling & Disabled & Disabled \\
Training schedule & 1 epoch; no explicit step cap & 1 epoch; no explicit step cap \\
Reported RLVR horizon & 260 steps (1.7B, 4B) & 200 steps (1.5B) \\
Save / evaluation interval & 20 / 20 steps & 20 / 20 steps \\
Evaluation temperature / top-$p$ & 0.6 / 0.95 & 0.6 / 0.95 \\
Evaluation samples per question & 4 & 4 \\
Student GPUs / rollout TP & 8 / 1 & 8 / 1 \\
Dynamic token budget per GPU & 32,768 & 32,768 \\
\bottomrule
\end{tabularx}
\end{table}

\subsection{Generalization Across Student Scale, Model Family, and Handoff Checkpoint}
\label{app:stage_ordering_generalization}

\begin{figure*}[!htbp]
\centering
\includegraphics[width=\textwidth]{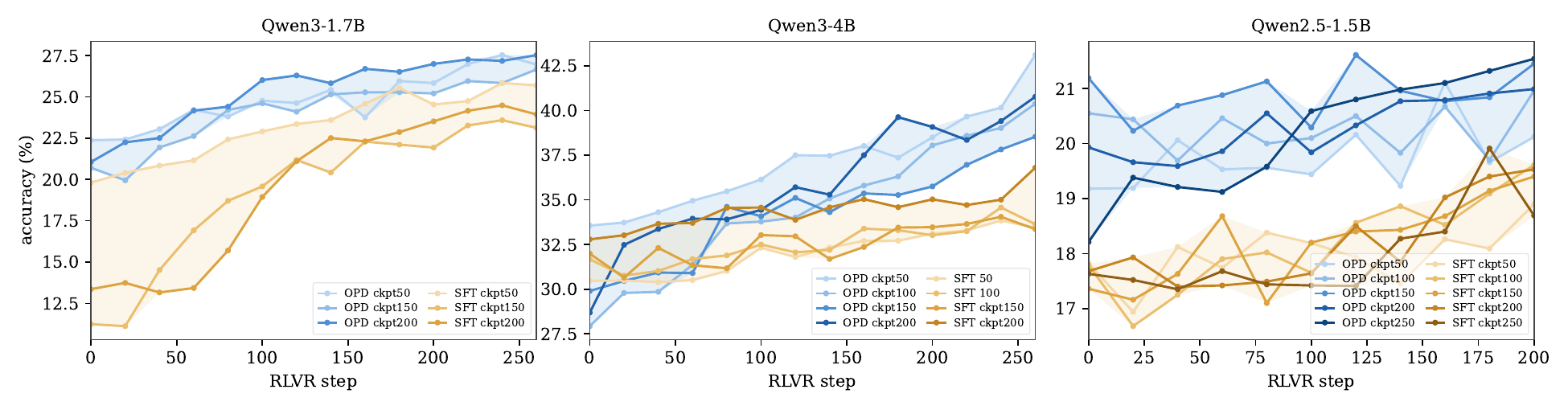}
\caption{
\textbf{The OPD initialization advantage reproduces across matched handoff checkpoints, student scales, and model families.}
Held-out \texttt{core-7} mean@4 accuracy for Qwen3-1.7B-Base (left),
Qwen3-4B-Base (center), and Qwen2.5-1.5B (right), using their corresponding 14B teachers.
Within each student family, OPD and SFT use the same downstream RLVR recipe and matched handoff-checkpoint grid.
Blue and orange curves denote OPD and SFT initialization.
Shading shows the minimum--maximum range across handoff checkpoints within each arm, not variation across random seeds.
The reported RLVR horizons are $260$ steps for Qwen3 and $200$ for Qwen2.5-1.5B.
}
\label{fig:stage_ordering_generalization}
\end{figure*}

We test whether the stage-ordering result in \S\ref{subsec:stage_ordering} generalizes beyond the main Qwen3-4B setting.
Figure~\ref{fig:stage_ordering_generalization} repeats the OPD-vs.\ SFT initialization comparison for Qwen3-1.7B and Qwen2.5-1.5B.
Within each student family, OPD and SFT are compared using the same student, teacher, RLVR data, RLVR recipe, and matched handoff-checkpoint grid.
Across every matched handoff setting, the OPD-initialized run finishes above its SFT-initialized counterpart.

The final accuracy ranges are $26.66$--$27.53\%$ versus $23.13$--$25.70\%$ for Qwen3-1.7B, $38.53$--$43.11\%$ versus $33.36$--$36.80\%$ for Qwen3-4B, and $20.12$--$21.54\%$ versus $18.69$--$19.61\%$ for Qwen2.5-1.5B.
Thus, the endpoint ordering is reproduced across all tested handoff checkpoints, two Qwen3 student scales, and the Qwen2.5 model family.

The strength of the control differs across settings.
For Qwen3-4B, OPD and SFT start from overlapping accuracy ranges and SFT is slightly stronger on average before RLVR, yet every OPD-initialized run finishes higher.
For the smaller students, OPD also begins RLVR at higher accuracy, so these experiments establish a consistent endpoint advantage but do not isolate the downstream RLVR gain from matched initial performance.
The checkpoint sweep provides repeated evidence across independently chosen handoff states, but it is not a substitute for variation across random training seeds.
We therefore use these experiments to support cross-model and cross-checkpoint reproducibility of the endpoint ordering rather than a claim of seed-level statistical significance.

\section{Case Study: Can Post-Training Composition Be Recursive?}
\label{app:iteration}

\begin{wrapfigure}{r}{0.34\textwidth}
\centering
\vspace{-15pt}
\includegraphics[width=\linewidth]{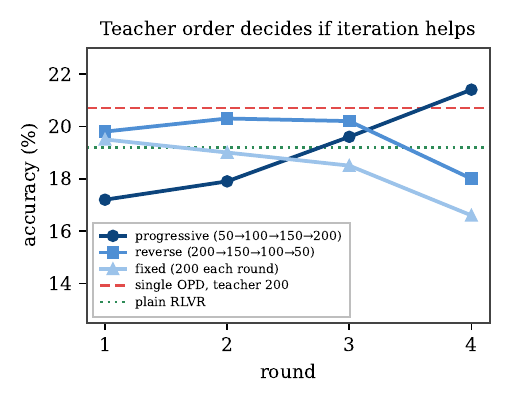}
\vspace{-20pt}
\caption{\textbf{Recursive composition under different teacher schedules.} Four SFT--OPD rounds on Qwen3-1.7B-Base, each using an RLVR-adapted Qwen3-14B checkpoint. Accuracy averages AIME~2024, AIME~2025, AMC~2023, and Minerva Math.}
\label{fig:iteration}
\vspace{-15pt}
\end{wrapfigure}

We ask whether repeated SFT--OPD composition remains productive as the student evolves, and how the teacher schedule affects this process.
Iterative self-training and self-play have been studied in STaR, ReST, and SPIN~\citep{zelikman2022star, gulcehre2023rest, chen2024spin}.
Here, we examine repeated composition of two stages while varying the teacher schedule.
The teacher checkpoints are prepared in advance; the student carries its learned state from one round to the next.

Starting from Qwen3-1.7B-Base, each round applies $200$ SFT steps on trajectories from one RLVR-adapted Qwen3-14B checkpoint, followed by a short OPD stage from the same teacher.
Handoffs use the observed OPD accuracy peak in each round.
We compare progressive ($t_{50}\!\rightarrow t_{100}\!\rightarrow t_{150}\!\rightarrow t_{200}$), reverse ($t_{200}\!\rightarrow t_{150}\!\rightarrow t_{100}\!\rightarrow t_{50}$), and fixed ($t_{200}$ in every round) schedules.
Progressive and reverse schedules reorder the same teacher checkpoints; the fixed schedule instead tests repeated use of the final teacher.
A single $200$-step OPD run from $t_{200}$ and a plain RLVR run serve as references.
The recursive schedules include additional SFT and OPD training, so total compute is not matched to these references.
\Cref{fig:iteration} reports accuracy averaged over AIME~2024, AIME~2025, AMC~2023, and Minerva Math.
Each schedule has one run.

\begin{wraptable}{r}{0.50\textwidth}
\vspace{-10pt}
\centering\small
\caption{Diagnostics recorded at the start of OPD and four-benchmark accuracy for every round using teacher $t_{200}$. Reinforceable fractions are shown on a $0$--$1$ scale.}
\label{tab:loop_dynamics}
\setlength{\tabcolsep}{4pt}
\begin{tabular}{lrrrr}
\toprule
Schedule & Round & rKL & $\rho^+$ & Acc. (\%) \\
\midrule
Fixed       & 1 & .42 & .48 & 19.5 \\
Fixed       & 2 & .39 & .46 & 19.0 \\
Fixed       & 3 & .38 & .45 & 18.5 \\
Fixed       & 4 & .38 & .44 & 16.6 \\
Reverse     & 1 & .43 & .47 & 19.8 \\
Progressive & 4 & .41 & .44 & 21.4 \\
\bottomrule
\end{tabular}
\vspace{-18pt}
\end{wraptable}

\textbf{Why this matters.} \quad
Teacher post-training produces a sequence of candidate supervision policies.
This permits a comparison between distilling directly from the final teacher and exposing the student to progressively more adapted checkpoints.
The case study asks whether the sequence of student--teacher pairings matters when SFT and OPD are repeated.

\paragraph{Discussion.}
Under the fixed schedule, the reinforceable fraction falls from $0.48$ to $0.44$, while four-benchmark accuracy decreases from $19.5\%$ to $16.6\%$ (\Cref{tab:loop_dynamics}).
Reverse KL also decreases, from $0.42$ to $0.38$, with little further change after round~2.
Repeated exposure to the final teacher therefore does not sustain improvement in this run.
The reinforceable fraction is also insufficient to explain differences across schedules: progressive and fixed round~4 both report $\rho^+=0.44$, yet their accuracies are $21.4\%$ and $16.6\%$, respectively.

The progressive schedule reaches the highest observed four-benchmark endpoint among the recursive runs; the reverse and fixed schedules do not sustain improvement across rounds (\Cref{fig:iteration}).
On the full seven-benchmark evaluation used in the main paper, however, the best recursive result remains $0.9$ percentage points below single-stage OPD.

Recursive runs reach the $8{,}192$-token response limit, truncating trajectories used for OPD; the single-stage OPD reference does not exhibit the same saturation.
This difference and the single run per schedule limit the interpretation of the observed ordering.
The case study motivates testing teacher schedules under matched training budgets and response-length conditions; it does not establish that recursive composition improves over a single OPD stage.


\section{Further Discussions}
\label{app:discussion}

\subsection{Why Composition Matters}
Our results suggest that post-training stages should be viewed as transformations of a \emph{learning state}, rather than independent sources of capability.
SFT trains on a fixed trajectory distribution, whereas RLVR and OPD train on states induced by the current policy.
Consequently, an earlier stage changes not only the model parameters but also the data distribution and supervision encountered by the next stage. This makes post-training updates naturally order-dependent, which we can express schematically as $\mathcal{U}_{B}(\mathcal{U}_{A}(\pi)) \neq \mathcal{U}_{A}(\mathcal{U}_{B}(\pi))$.

Our experiments expose three manifestations of this non-commutativity.
First, RLVR can produce a substantially stronger student while making subsequent OPD destructive, showing that capability and distillability are distinct.
Second, changing the teacher through RLVR substantially changes what can be transferred through an otherwise identical OPD stage.
Third, accuracy-matched SFT and OPD checkpoints respond very differently to subsequent RLVR, even when their aggregate entropies overlap.
Together, these results motivate our central principle: a post-training stage should be evaluated not only by the capability it adds, but also by the learning interface it leaves for the next stage.

\subsection{Limitations}
Our conclusions are deliberately narrower than a universal post-training recipe:

\begin{itemize}[leftmargin=12pt,itemsep=2pt]
\item \textbf{Composition is tested more narrowly than OPD scaling.}
Our OPD diagnostics span multiple student and teacher scales, architectures, objectives, and datasets, whereas the main composition study centers on a Qwen3-4B student and Qwen3-14B teacher.
We therefore claim a controlled demonstration of stage interaction, not that the same ordering is optimal for every model or domain.

\item \textbf{Some controls separate practical procedures rather than causal mechanisms.}
In particular, RLVR substantially improves the teacher while our STaR-style control does not.
This establishes that RLVR is the more effective teacher-preparation method under our protocol and that its added capability transfers through OPD; it does not establish that RLVR-induced capability is intrinsically more transferable at matched teacher accuracy.

\item \textbf{Our diagnostics are predictive, not sufficient explanations.}
$\rho_0^+$ separates productive and destructive student--teacher states where scalar reverse KL does not, and entropy alone does not explain the OPD-to-RLVR advantage.
These results rule out simple scalar explanations, but they do not identify the complete mechanism governing downstream learnability.

\end{itemize}

These limitations do not change the primary conclusion: across controlled interventions, \emph{the same stage behaves differently depending on the state created by the stage before it}.

\subsection{Unsuccessful Attempts}
Several negative results were particularly informative because they rule out seemingly natural alternatives.

\begin{itemize}[leftmargin=12pt,itemsep=2pt]
\item \textbf{Maximizing pre-OPD student capability.}
RLVR produces a much stronger student than a short SFT warm-up, yet subsequent OPD can regress it below its starting accuracy.
Thus, selecting a warm-up solely by pre-OPD accuracy can choose the worse downstream initialization.

\item \textbf{Training each stage longer.}
Additional SFT eventually stops improving downstream OPD, while continued OPD can reduce task accuracy even as teacher--student KL continues to improve.
More optimization of an intermediate objective is therefore not necessarily better optimization of the complete pipeline.

\item \textbf{Reusing the strongest teacher recursively.}
In our recursive case study, repeatedly using the final teacher does not continue to improve the student, while progressive teacher refresh is the only tested ordering that improves across rounds.
Because these runs are not fully replicated and exhibit trajectory-length truncation, we interpret this as evidence that supervision can become stale as the student changes, rather than as proof of an optimal recursive algorithm.

\end{itemize}

These failures reinforce the same pattern as the positive results: locally improving one stage can worsen the conditions faced by the next.

\end{document}